\documentclass[runningheads]{llncs}
\usepackage{graphicx}
\usepackage{comment}
\usepackage{amsmath,amssymb} 
\usepackage{color}
\usepackage{xcolor}

\usepackage[numbers]{natbib}
\usepackage[width=122mm,left=12mm,paperwidth=146mm,height=193mm,top=12mm,paperheight=217mm]{geometry}
\begin{document}
\pagestyle{headings}

\title{Visual Attention in Imaginative Agents} 

\author{Samrudhdhi B. Rangrej, 
James J. Clark}
\authorrunning{S. Rangrej et al.}
\institute{Centre for Intelligent Machines \& ECE Department\\McGill University, Montreal, QC, Canada}

\maketitle
\begin{abstract}
We present a recurrent agent who perceives surroundings through a series of discrete fixations. At each timestep, the agent imagines a variety of plausible scenes consistent with the fixation history. The next fixation is planned using uncertainty in the content of the imagined scenes. As time progresses, the agent becomes more certain about the content of the surrounding, and the variety in the imagined scenes reduces. The agent is built using a variational autoencoder and normalizing flows, and trained in an unsupervised manner on a proxy task of scene-reconstruction. The latent representations of the imagined scenes are found to be useful for performing pixel-level and scene-level tasks by higher-order modules. The agent is tested on various 2D and 3D datasets.
\end{abstract}
\section{Introduction}
Attention is central to visual perception. When an agent is performing a task, it is imprudent to dedicate resources for the detailed sensing of an environment. Actively attending to the limited but most informative regions of an environment is crucial for efficiency. An intelligent agent is one which can make use of a few informative views to determine an entire surrounding. 

Humans analyze scenes by paying attention to various areas of the visual field one after another. The eyes successively fixate on discrete locations of a scene and move very quickly between consecutive fixations. The details captured from various fixations are progressively fused to build a mental representation of a scene \cite{spatiotopic}, which is subsequently used for higher-order tasks.

To act confidently, an agent has to know its surroundings accurately. Complete certainty may be achieved if an agent scans an entire scene. However, time-critical tasks demand to learn about the surroundings without waiting for the scanning to finish. But due to the regularities found in the world, it seems that it should be possible to predict the surroundings from a limited set of views.

Humans can `guess' the surroundings from obstructed views \cite{priming}. Based on a narrow view from a window showing a few trees, we can imagine a forest or a garden outside. Similarly, from a limited interior view containing bookshelves, we can imagine tables and chairs in the surroundings and can predict that it is a scene of a study room or a library. Gathering information from a visual scene is a recurrent process. Information gathered so far helps in hypothesizing the content from as yet unseen regions. One can then shift attention to a region that is deemed important based on these hypotheses. As the sensing progresses, the hypotheses are refined. One can predict the entire scene without attending to unimportant regions.

Hypothesizing plausible surroundings has several benefits. It helps in reducing errors while making decisions with partial observations. A partially observed scene may not contain the critical features required for making a decision. Furthermore, it may contain distracting information resulting in the wrong decision. Consider the task of `cat' vs. `dog' classification and consider a test image of a dog containing a toy shaped like a cat. If a glimpse is on the toy, the agent may wrongly classify the image as `cat'. However, if the agent can hypothesize a dog in the surrounding, it may choose to gather more evidence before making a decision. 

Hypothesizing plausible surroundings also facilitates planning. An agent can focus on likely scenarios while eliminating unlikely ones. While driving, if the agent sees a house and hypothesizes that it is entering a residential territory as opposed to a freeway, it can prepare to brake often. Similarly, if a robotic agent can hypothesize a 3D structure of an object from a few strategic 2D views, it can plan grasping and object manipulation. Furthermore, these types of agents are more trustworthy compared to black-box agents as they reveal the imagined hypotheses on which the decisions are based. Also, hypothesizing complete surroundings from the past set of glimpses resists forgetting.

We propose to build an imaginative agent which imagines the yet unseen portions of the scene. Higher-order modules can use the imagination for making task-specific decisions. Imagining surroundings from a limited number of glimpses is an ill-posed problem. One can imagine several surrounding scenes that are consistent with the seen glimpses. If an agent is optimized to make a single prediction, it predicts a blurry scene - a numerical average of all possibilities. Hence, we will allow our agents to predict many plausible surroundings.

A problem of how to capture informative views remains. A naive approach is to scan the scene in a sliding-window fashion, which is sub-optimal. An optimal next view should be the one which maximally reduces the uncertainty in the prediction of the surrounding. An agent should start with fixating on those locations that help in learning the most about the unseen surrounding and gradually move to the less informative locations. Informativeness of a view can be decided based on its contribution to imagining the unseen surrounding. A view is deemed informative if it reduces the uncertainty in the imagined content. The eventual low uncertainty in the imagination obviates the necessity of further fixations. This can improve the performance of an agent in time-critical tasks \cite{bajcsy1988active}.

\subsection{Problem Description}

We address two related problems of hypothesizing surrounding scenes from limited views and deciding the next most informative view based on the predicted surroundings. An agent captures a partial view of a scene and hypothesizes a complete scene. Uncertainty due to the ill-posedness of the problem is tackled by allowing the agent to make multiple simultaneous predictions. The predictions are pseudo-realistic and interpretable. The agent maintains a set of internal representations, each corresponding to a different hypothesis. As new views become available, the representations are filtered to exclude false hypotheses. The agent is trained in an unsupervised fashion. Unsupervised training makes use of unlabelled data, which is available in abundance and at a low cost. Once the training is finished, the imagined scenes or their internal representations can be used for higher-order tasks.

The design of our agent is inspired by the biological phenomena of maintaining a visual memory and guessing surroundings from glimpses based on memory. The human brain compresses observed visual scenes into latent codes to remember them efficiently. The compression retains only high-level information \cite{changeblindness}. Semantically similar visual concepts are encoded with similar codes \cite{perceptualgrouping}. The high-level information can be recalled by decoding the latent codes. Additionally, the same latent codes are used for higher-order vision tasks. The brain also possesses the ability to imagine new scenes that are similar to past visual experiences \cite{remembering}. Humans can also imagine the surroundings given partial observation(s) of a scene \cite{priming}. The imagination is derived from visual memory. It is easy to guess the surrounding if similar scenes are present in the visual memory. 

Our agent is based on a variational autoencoder, which recurrently generates complete scenes conditioned on a sequence of partial observations. Scenes generated by VAEs are associated with a distribution of latent representations. To use the representations for a scene-level task, they should contain the information of the entire scene independent of the history of the partial observations. We develop a novel technique to warp and unwarp the latent space of an unconditional VAE based on the observed views. Two normalizing flows learn the warping and unwarping functions conditioned on the observation history. During the prediction phase, samples from the warped latent space are mapped to the unconditional latent space using a conditional unwarping function. The unconditional decoder decodes the mapped samples to generate complete scenes. Lastly, we demonstrate that the unconditional representations can be used for higher-order scene-level tasks.

The next fixation is planned based on the uncertainty in the predicted scenes. A next view is captured on the location where the uncertainty is highest. Sensing in this manner reduces uncertainty in the subsequent predictions.

\section{Literature Review}

\subsubsection{Visual Attention.} 
The computer vision community has used attention to solve many vision tasks. Early traces of visual attention can be found in multi-scale edge detection algorithms \cite{kellyfaceedge}. \citet{neuralgnostic} proposed neural networks that attend to and recognize patterns in an image sequentially. \citet{kochullman} proposed a winner-take-all algorithm for visual attention. Later, \citet{ittisaliency} extended the method by including image saliency. \citet{clarkmodal} tuned the weights of salient features based on feedback. \citet{colorhistfix} suggested paying attention in the direction of the gradient of the object matching function. \citet{histbackproj} developed a method called histogram back-projection to identify the important regions to attend. \citet{origami} planned fixations on the prominent edges to recognize origami objects.

Several researchers used graphical models to describe visual attention. \citet{ahmm} modeled eye-movements using an augmented Hidden Markov Model. \citet{ipomdp} learned eye-movement for visual search task by minimizing the entropy of a belief state used for predicting the location of a target. \citet{fastod} extended the model for multiple targets. \citet{fovealrbm} used third-order connections in Restricted Boltzmann Machine(RBM) to learn a relationship between the context, the location of fixations, and the internal representation of a scene. Furthermore, an approach based on an autoregressive model was proposed by \citet{fixationnade}. \citet{searchingfromcontext} proposed a learning-free method based on nearest neighbors. 

Recently, the machine learning community developed `soft' and `hard' visual attention. Soft attention allows the content of an entire scene to be perceived at once, but with varying degrees of importance. A deterministic attention map is predicted and multiplied with the input to suppress unimportant content \cite{showattendtell, draw}. Soft attention has been used for solving various computer vision tasks such as image classification and segmentation \cite{softclassification, scaleattention}, image captioning \cite{showattendtell}, optical character recognition \cite{softocr}, and pose estimation and action recognition \cite{softposeestimation, softactionrecognition}. Hard attention refers to a mechanism where an agent perceives a scene through a series of fixations. The agent decides the location of the next glimpse stochastically \cite{ram}. Hard attention has been employed for solving various computer vision tasks such as image captioning \cite{showattendtell}, object recognition and localization \cite{multiobject, localandreco, activelocalization}, and 3D scene classification  \cite{lookahead}. The REINFORCE optimization algorithm used for learning hard attention has high variance and is very sensitive to the hyper-parameter setting. As a solution, \citet{emfovea} proposed optimization using an EM-like algorithm, and \citet{wakesleep} proposed a wake-sleep algorithm.

Active vision is closely related to hard attention. An active agent has more degree of freedom to evaluate a scene from multiple viewpoints. A large body of work is focused on actively perceiving a 3D environment for scene recognition and localization  \cite{activelocalization}, and manipulating a 3D object for recognition \cite{activevisiondataset}. Another line of work addresses active vision-based view-planning \cite{viewplan1}, navigation and localization \cite{slam}. Fast and timely agents start recognizing objects, though imperfectly, from the very first viewpoint and update the decision as time progresses  \cite{timelyrecognition}.

Similar to the paradigms of hard attention and active vision, our agent perceives the surrounding through a series of glimpses. However, learning is not dedicated to any supervised task. The agent learns to explore the surroundings by solving an unsupervised proxy task of constructing complete scenes from partial observations.

\subsubsection{Prediction of the Surrounding.}
The task of predicting surroundings is similar to the tasks of image in-painting \cite{inpainting1}, context prediction \cite{tellmewhereiam}, view-synthesis  \cite{neuralrender}, predicting 3D structure from 2D views \cite{2dto3d1}, and predicting missing pixels \cite{svae}. These are passive tasks where an active exploration of scenes is not performed. Our agent explores the scene actively while learning from a proxy task of scene-construction.

\subsubsection{Proxy Tasks.}

In computer vision, proxy tasks are often used to improve performance on the main task. Examples are colorization for visual understanding \cite{colorization}, learning depth maps by reconstructing stereo images \cite{proxydepth}, crowd counting by learning to rank \cite{crowdleveraging}, cross channel prediction \cite{splitbrain}, and feature learning by predicting image rotation \cite{unsupervisedrotation}, by solving jigsaw puzzle \cite{jigsaw} or by learning to count visual primitives \cite{countpremitive}, etc.

\subsubsection{Prediction of Multiple Hypotheses.}

Prediction of surroundings from a small set of glimpses is an ill-posed problem as there exist several non-unique solutions. A model might learn to predict an average of all possible solutions if the uncertainty is not accounted for \cite{finnvideo}. A widely used approach is to find a fixed number of multiple simultaneous solutions. The method is used for several ill-posed tasks such as future frame-prediction \cite{mult1}, optical flow estimation \cite{mult3}, classification, segmentation and caption generation \cite{mult4}. A variational approach overcomes the constraint of predicting a limited number of hypotheses. The approach allows unlimited sampling of hypotheses from a learnt distribution. The method has been used to solve several ill-posed tasks such as segmentation \cite{mult2} and motion prediction \cite{mult5}.
\bigbreak
Our problem is similar to the ones addressed in \cite{lookaround,lookaround2,lookaround4}. They also use scene reconstruction as a proxy task for learning visual attention. However, their agents predict a single scene from partial observations. Additionally, they learn policies for scene exploration, whereas our approach uses a predefined policy.

\section{Background}
The proposed imaginative agent is built using variational autoencoders(VAE) and normalizing flows(NF). A brief review of these is provided next.

\subsection{Variational Autoencoder}
A variational autoencoder(VAE) is a latent-variable model that assumes that the data is generated from a low dimensional latent representation. The data $x$ and the latent representation $z$ are generated from the likelihood model $p(x|z)$ and the prior $p(z)$, respectively. Learning the likelihood model is challenging as the posterior $p(z|x)$ is intractable. Therefore, a surrogate posterior $q(z|x)$ is inferred. The posterior $q(z|x)$ and the likelihood $p(x|z)$ are modeled using an encoder and a decoder network, respectively. Network parameters are optimized by maximizing the evidence lower bound (ELBO) objective \cite{vae}.
\begin{align}
    L_{VAE} = \mathbb{E}_{z \sim q(z|x)} \log p(x|z) - \mathbb{KL} (q(z|x) || p(z))
    \label{eq:vae}
\end{align}

The expectation in the above formula is approximated using samples $z$ drawn from $q(z|x)$. Despite sampling, differentiability is preserved using the reparameterization trick. The prior $p(z)$ is assumed to be a multivariate Gaussian with unit variance and $q(z|x)$ is modeled as a factorized Gaussian $\mathcal{N}(\mu(x), \Sigma(x))$.

The ELBO in Eq. \ref{eq:vae} is tight when $q(z|x) \approx p(z|x)$. However, the factorized Gaussian assumption limits the capacity of the posterior in making quality inferences \cite{inferencegaps}. Alternate approaches are proposed to increase the flexibility of the posterior such as using a mixture of Gaussians \cite{gammprior1} and normalizing flows \cite{flowvae} to model the posterior, building hierarchical latent spaces \cite{laddervae} and using importance weighted sampling \cite{ImportanceWA}. We use normalizing flow, a scalable tool, to learn a flexible and arbitrarily complex posterior $q(z|x)$ that fits the true posterior $p(z|x)$ more closely.

\subsection{Normalizing Flow}
Normalizing flow (NF) maps a primary posterior $q(z|x) \equiv q(z_0|x)$ to an alternate posterior $q(z_K|x)$ using a series of $K$ invertible transformations. A single invertible map $f_1$ transforms a primary latent representation $z_0$ into an alternate latent representation $z_1$. According to the change of variable formula,
\begin{align}
    q(z_1|x) &= q(z_0|x) |\text{det}(J_{f_1})|^{-1} \label{eq:flow}
\end{align}
Here, $J_{f_1}$ is the Jacobian of $f_1$. The normalizing flow transforms a factorized Gaussian posterior $q(z_0|x)$ to a more complex posterior $q(z_K|x)$ using a composition of several invertible maps $f = f_K \circ \dots \circ f_2 \circ f_1$. The final posterior can be derived by repeated application of Eq. \ref{eq:flow}.
\begin{align}
    q(z_K|x) &= q(z_0|x) \prod_{\kappa=1}^{K}|\text{det}(J_{f_\kappa})|^{-1}
\end{align}
The above posterior replaces the primary posterior in the ELBO objective defined in Eq. \ref{eq:vae}. The law of the unconscious statistician (LOTUS) \cite{inferencegaps} simplifies the computation of the expectation.
\begin{align}
    L_{VAE} &= \mathbb{E}_{z_0 \sim q(z_0|x)} \log p(x|z_K) - \mathbb{KL} (q(z_K|x) || p(z_K)) 
    \label{eq:flowvae}
\end{align}
Examples of neural architectures for normalizing flow are IAF \cite{iaf}, MAF \cite{maf}, NICE \cite{NICE}, Real NVP \cite{RNVP}, Glow \cite{glow}, NAF \cite{NAF} and BNAF \cite{BNAF}. We adapt Block Neural Autoregressive Flow (BNAF) in our implementation. 

\subsection{Conditional Variational Autoencoder}
A conditional variational autoencoder(cVAE) \cite{cvae} generates data $x$ from a latent variable $z$ based on an input observation $h$. Similar to what we did for unconditional VAEs, a NF can be used here as well to model a complex posterior for more accurate inference. Eq. \ref{eq:flowvae} can be modified to define the conditional ELBO objective as follows.

\begin{align}
    L_{cVAE} = \mathbb{E}_{z \sim q(z_0|x, h)} \log p(x|z_k, h) - \mathbb{KL} (q(z_k|x, h) || p(z_k|h))
    \label{eq:cvae}
\end{align}

A conditional prior $p(z_k|h)$ is assumed to be a multivariate Gaussian with unit variance. The posterior $q(z_0|x,h)$ is transformed to $q(z_k|x,h)$ using NF. 

\begin{figure}[!t]
    \centering
    \vfill
    \begin{minipage}[c]{\linewidth}
    \includegraphics[width = \textwidth]{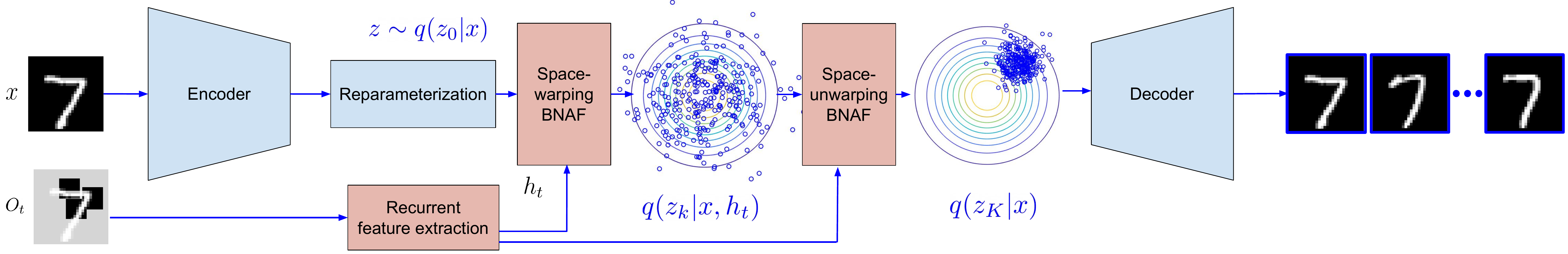}
    \centering{(a)}
    \end{minipage}
    \vfill
    \hfill
    \begin{minipage}[c]{0.55\linewidth}
    \includegraphics[width = \textwidth]{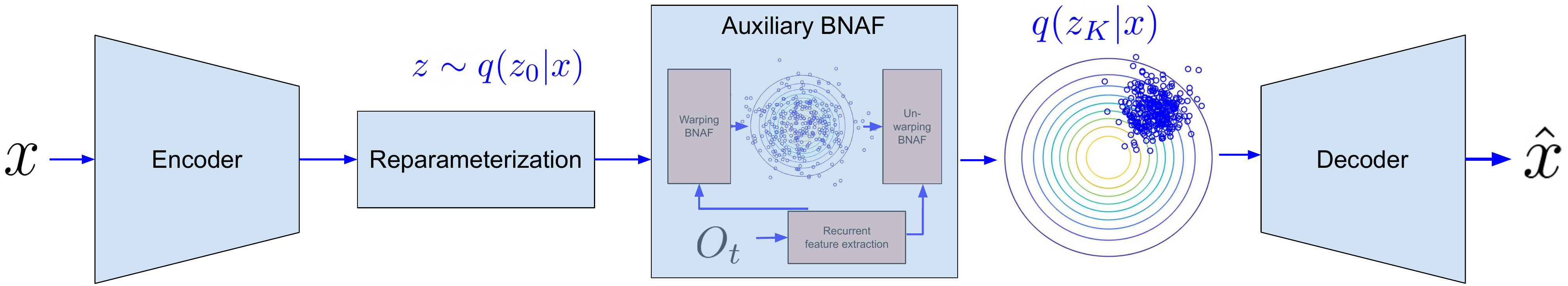}
    \centering{(b)}
    \end{minipage}
    \hfill
    \begin{minipage}[c]{0.43\linewidth}
    \includegraphics[width = \textwidth]{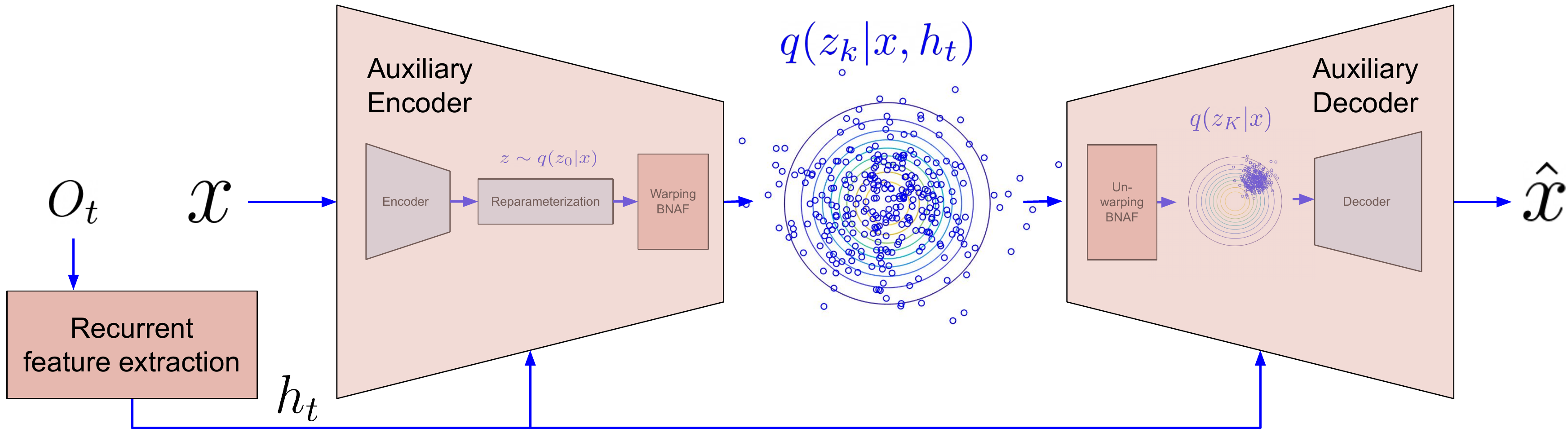}
    \centering{(c)}
    \end{minipage}
    \hfill
    \vfill
    \centering
    \begin{minipage}[c]{\linewidth}
    \includegraphics[width = 0.75 \textwidth]{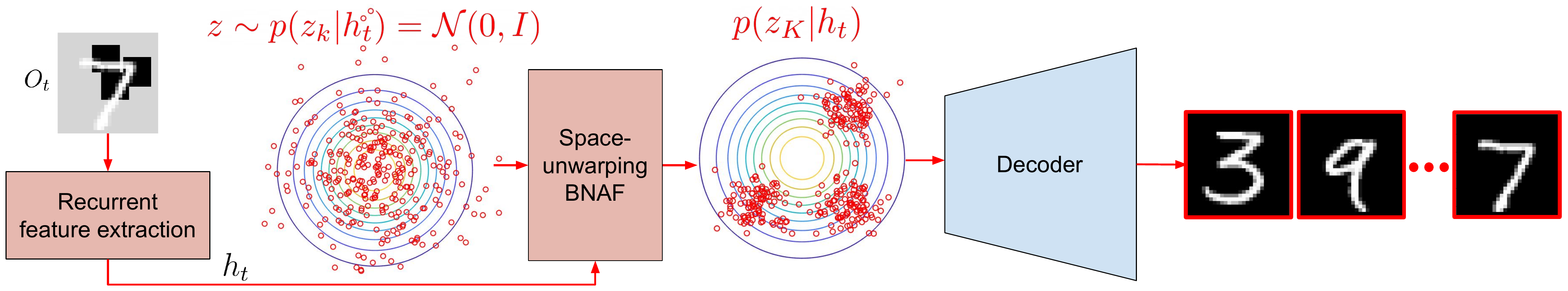}
    \centering{(d)}
    \end{minipage}
    \vfill
    \caption{The proposed model. Unconditional and conditional modules are shown in blue and red respectively. (a) Training phase. The encoder infers a latent distribution $q(z_0|x)$ from an input $x$. The first \textit{BNAF} warps $q(z_0|x)$ to $q(z_k|x, h_t) \approx \mathcal{N}(0,I)$ based on features $h_t$ extracted from observation history $O_t$. The second \textit{BNAF} unwarps $q(z_k|x, h_t)$ to $q(z_K|x)$. The decoder reconstructs $x$ from $q(z_K|x)$. (b) The model viewed as an unconditional VAE. The space-warping and unwarping BNAFs are grouped to form an auxiliary BNAF. (c) The model viewed as a conditional VAE. The encoder and the space-warping BNAF are grouped as an auxiliary encoder. The space-unwarping BNAF and the decoder are grouped as an auxiliary decoder. (d) Prediction phase. An auxiliary decoder is used to predict scenes from partial observations. The second \textit{BNAF} unwarps $\mathcal{N}(0,I)$ to $p(z_K|h_t)$. The posterior $p(z_K|h_t)$ is a multi-modal distribution. Each mode gives rise to a scene with different global structure when decoded using the decoder. The constructed scenes are consistent with $O_t$.\vspace*{-0.7cm}}
    \label{fig:model}
\end{figure}

\section{Imaginative Agent}

A recurrent agent sequentially observes glimpses covering small portions of a scene and hypothesizes complete scenes that are coherent with the partial observations. The model has four main building blocks as shown in Fig \ref{fig:model}(a). 

\textbf{Unconditional encoder and decoder.}
An encoder network infers a posterior $q(z_0|x)$ from a complete scene $x$. Then, a sample $z_0 \sim q(z_0|x)$ is transformed to $z_K$ using BNAFs, as discussed next. A decoder network reconstructs $x$ from $z_K$. Note that the encoder and the decoder are unconditional. They are independent of the sensing process.

\textbf{Sensor.} A sensing function $s_w(x, l)$ extracts a square glimpse (space-limited sample of the scene) of size $w \times w$ centered at location $l$ from a given scene $x$. Let the observations until time $t$ be defined as $O_t(x) = \{(s_w(x, l_\tau), l_\tau) | \tau \in {1, 2, \dots, t}\}$.

\textbf{Recurrent feature extractor.}
A recurrent neural network(RNN) extracts fixed-length features from observations $O_t(x)$. At time $\tau$, the RNN updates its hidden state by processing and incorporating features extracted from a tuple $(s_w(x, l_\tau), l_\tau)$. The final hidden state after $t$ timesteps is output as the feature map $h_t$.

\textbf{Conditional BNAFs.}
The original unconditional BNAF presented in \cite{BNAF} is made conditional by connecting each hidden layer to an additional input $h_t$. Conditional BNAFs learn to warp and unwarp a latent space between the pair of unconditional encoder and decoder. The first BNAF warps an unconditional posterior $q(z_0|x)$ to a Gaussian $\mathcal{N}(0,I)$, and the second BNAF unwarps $\mathcal{N}(0,I)$ to an unconditional posterior $q(z_K|x)$.

\subsection{Training Objective}
The building blocks can be grouped to view the proposed system as either unconditional or conditional VAEs. Grouping under the unconditional view is shown in Fig \ref{fig:model}(b), where a combination of the two BNAFs can be seen as a single auxiliary BNAF. The encoder learns to make an inference from a complete scene, and the decoder reconstructs the same scene. The auxiliary BNAF learns a flexible posterior $q(z_K|x)$ while ignoring $O_t$ as noise. In this setting, the system can be trained by optimizing the $L_{VAE}$ objective defined in Eq. \ref{eq:flowvae}.

Grouping under the conditional view is shown in Fig \ref{fig:model}(c). A pair of encoder and space-warping BNAF is identified as an auxiliary encoder. Similarly, a pair of space-unwarping BNAF and decoder is identified as an auxiliary decoder. The auxiliary encoder and decoder are conditioned on $h_t$. The auxiliary encoder performs inference on a complete scene $x$ and features $h_t$. The auxiliary decoder reconstructs $x$ from a latent representation $z_k$ and features $h_t$. The model can be optimized by using the $L_{cVAE}$ objective defined in Eq. \ref{eq:cvae}.

The agent is trained by optimizing a linear combination of the unconditional and the conditional objectives. In practice, better performance is achieved by modifying the conditional objective using a trick used in \cite{cvae}. The idea is to train auxiliary encoder and decoder using separate training objectives. The auxiliary encoder is trained to minimize $\mathbb{KL} (q(z_k|x, h_t) || p(z_k|h_t))$ without the help of the auxiliary decoder, i.e. it does not receive any gradient from $\log p(x|z_k, h_t)$. Similarly, the auxiliary decoder should be trained to predict the entire scene $x$ conditioned on $h_t$ without the help of the remaining components. However, direct optimization of $\log p(x|h_t)$ is not possible as the posterior $p(z_k|x, h_t)$ is not known to the auxiliary decoder. Instead, we train the auxiliary decoder to reconstruct locally the portion of the scene observed by the agent so far. This is done by optimizing ELBO on masked $x$. The mask $m_t(\cdot)$ segregates pixels observed until time $t$. The ELBO on $\log p(m_t(x))$ is given below.
\begin{align}
    \log p(m_t(x)) & \leq \mathbb{E}_{p(z_k|h_t)} \eta  \log p(m_t(x)|z_K) - \mathbb{KL} (p(z_K|h_t) || p(z_K)) 
\end{align}

The coefficient $\eta$ is the ratio of the number of pixels in $x$ to the number of pixels retained by $m_t(\cdot)$. The final training objective at time $t$ is as follows.
\begin{align}
    L_t &= \mathbb{E}_{q(z_0|x)} \log p(x|z_K) - \mathbb{KL} (q(z_K|x) || p(z_K)) - \mathbb{KL} (q(z_k|x, h_t) || p(z_k|h_t)) \nonumber \\
    &+ \mathbb{E}_{p(z_k|h_t)} \eta  \log p(m_t(x)|z_K) - \mathbb{KL} (p(z_K|h_t) || p(z_K)) 
    \label{eq:Lfinal}
\end{align}

\begin{figure}[!t]
    \centering
    \begin{minipage}[c]{0.13\linewidth}
    \includegraphics[width = \textwidth]{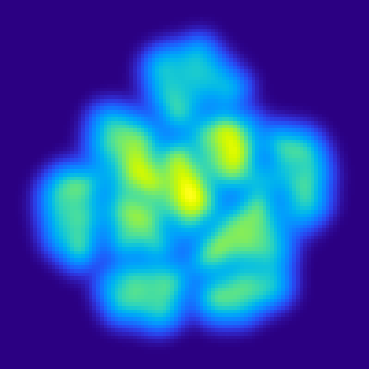}
    \centering{(a)}
    \end{minipage}
    \hfill
    \begin{minipage}[c]{0.13\linewidth}
    \includegraphics[width = \textwidth]{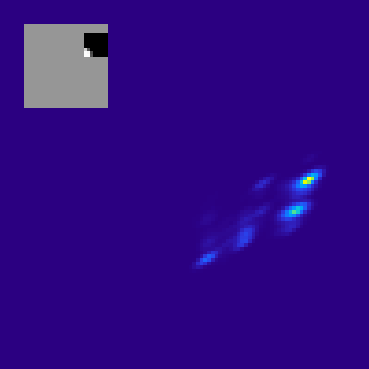}
    \centering{(b)}
    \end{minipage}
    \hfill
    \begin{minipage}[c]{0.13\linewidth}
    \includegraphics[width = \textwidth]{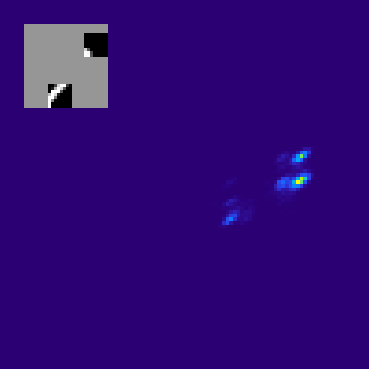}
    \centering{(c)}
    \end{minipage}
    \hfill
    \begin{minipage}[c]{0.13\linewidth}
    \includegraphics[width = \textwidth]{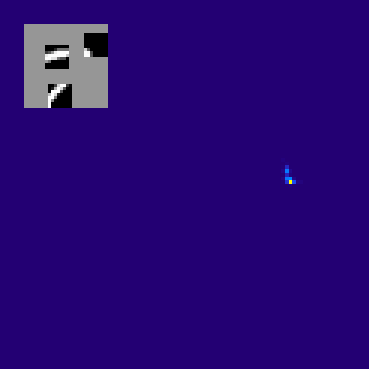}
    \centering{(d)}
    \end{minipage}
    \hfill
    \begin{minipage}[c]{0.13\linewidth}
    \includegraphics[width = \textwidth]{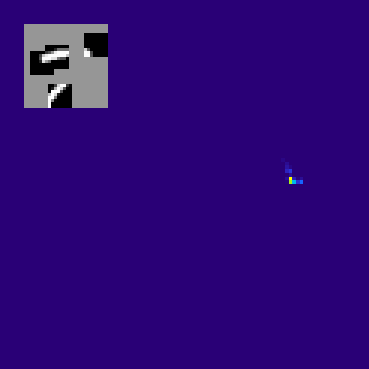}
    \centering{(e)}
    \end{minipage}
    \hfill
    \begin{minipage}[c]{0.13\linewidth}
    \includegraphics[width = \textwidth]{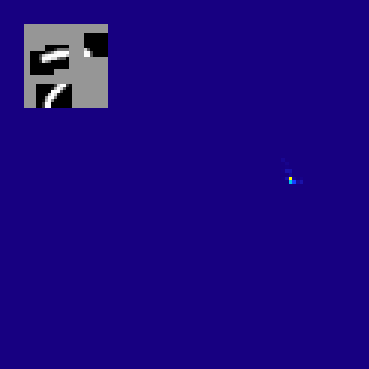}
    \centering{(f)}
    \end{minipage}
    \caption{(a) A T-SNE projection \cite{tsne} of unconditional latent space $q(z_K)$ for MNIST dataset. (b-f) T-SNE projections of unwarped posteriors $p(z_K|h_t)$ for MNIST image from t=1 to 5. Partial observations are shown in the upper left corner.\vspace*{-0.0cm}}
    \label{fig:unwarpedlatent}
\end{figure}

The complete model is optimized by maximizing $\sum_{t} L_t$. The first two terms in  Eq. \ref{eq:Lfinal} help in creating visual memory by learning compact representations of the complete scenes. The third term arranges $z_k$ corresponding to different scenes with a similar $O_t$ to follow a Gaussian density. The Gaussian density unwarps to a multi-modal posterior $p(z_K|h_t)$ during the prediction phase (see Fig.\ref{fig:unwarpedlatent}(b)). Each mode may generate scenes with any realistic global structure, but the content on the observed locations should be consistent with $O_t$. The last two terms ensure local consistency and global realism, respectively.

\subsection{Prediction Phase}
Prediction of complete scenes from partial observations only requires the auxiliary decoder. A detailed view of the auxiliary decoder is shown in Fig.\ref{fig:model}(d). Note that the complete scene $x$ is not available to the agent during the prediction phase. The agent extracts features $h_t$ from a set of partial observations $O_t$. The features $h_t$ are used by the BNAF to perform the conditional unwarping of a Gaussian latent space. Multiple samples from the prior $p(z_k|h_t) = \mathcal{N}(0,I)$ are unwarped to the representations $z_K$s simultaneously. The decoder decodes the $z_K$s and generates realistic scenes that have different global structures but are consistent with the partial observations. As shown in Fig.\ref{fig:model}(d), when an agent observes a portion of a scene containing an angular stroke on the upper right corner, it generates digits `3', `9', and `7', which have similar angular strokes.

\subsection{Uncertainty-based Sensing}
\label{sec:sense}
We can quantify the variety in the scenes predicted at time $t$ using a pixel-wise variance-map $V_t$. The conditional BNAF and the decoder are represented as $f_{h_t}$ and $D$. A mean scene is defined as $\mu_t$.
\begin{align}
      V_t &= \int ||D(z_K)- \mu_t||^2 p(z_K|h_t) dz_K = \mathbb{E}_{z_k \sim p(z_k|h_t)} ||D\big(f_{h_t}(z_k)\big)- \mu_t||^2 
    \label{eq:uncertain}
\end{align}

The variety in the predicted scenes is high initially, as there are many possible scenes with the same $O_t$. As time passes, $O_t$ covers a larger area of the scene, and the variety in the possible distinct predictions reduces. When all structure defining parts of the scene are observed, the predicted scene is the same as the true scene. This is also evident from Fig.\ref{fig:unwarpedlatent}. The unwarped posterior at $t=1$ is multi-modal. Each mode corresponds to a different hand-written digit. As the agent observes more regions of the scene, it eliminates inconsistent possibilities, and the number of modes in $p(z_K|h_t)$ reduces. Eventually, $p(z_K|h_t)$ becomes uni-modal, representing a single possibility corresponding to the actual scene.

In most time-critical tasks, it is crucial to predict the actual surrounding from a limited number of observations. This is possible if the consecutive observations help in eliminating a maximum number of wrong hypotheses. After the first random fixation, the consecutive fixations should sit on the location where $V_t$ is highest since $V_t$ represents uncertainty in the content predicted by the agent. Sensing on the locations where the uncertainty is highest achieves two goals. First, the actual content on the most uncertain location can be determined. Second, uncertainty in the content of the other locations can be reduced due to structural dependence found in the natural scenes. The observations in Fig.\ref{fig:unwarpedlatent} are captured using uncertainty-based sensing. The agent achieves certainty from 3 observations covering less than 25\% of the total area.

\section{Experiments}
The proposed system is validated on several 2D and 3D datasets. Refer to the supplementary material for details about the implementation.

\noindent\textbf{MNIST \cite{mnist}.} The dataset consists of gray-scale images of hand-written digits of size $28\times28$. The train and test sets contain 60K and 10K images, respectively. An agent is trained for 7 time-steps using the observations of size $8\times8$.

\noindent\textbf{SVHN \cite{svhn}.} The dataset is a collection of color images of street-view house numbers of size $32 \times 32$. There are approximately 0.6M train and 26K test images. An agent is trained on the observations of size $11 \times 11$ for 8 time-steps.

\noindent\textbf{ModelNet40 \cite{modelnet}.} The dataset is a collection of CAD models of 3D objects belonging to 40 different categories. The train and test sets contain approximately 10K and 2.5K objects. A 2D grid of 7$\times$12 views is generated by capturing views of size $32\times32$ from 7 camera elevations $(0^\circ$, $30^\circ$, $60^\circ$, $90^\circ$, $120^\circ$, $150^\circ$, $180^\circ)$ and 12 azimuths $(0^\circ$, $\pm 30^\circ$, $\pm60^\circ$, $\pm90^\circ$, $\pm120^\circ$, $\pm150^\circ$, $180^\circ)$. An agent is trained to predict the entire view-grid from up to 8 2D views.

\noindent\textbf{Cityscapes \cite{cityscapes} + BDD \cite{bdd}} These datasets provide urban scene images captured by vehicle-mounted cameras. Train sets of the two datasets are combined to make a large train set of approximately 10K images. The validation set of 1.5K images from the Cityscape dataset is used as a test set. The images are downsampled to the size of $320\times640$ prior to the training. An agent is trained on the observations of size $160\times160$ for 6 timesteps. The agent directly predicts segmentation maps of the surroundings instead of the visual images. This requires supervision from human generated ground truth.

\begin{figure}[!t]
    \vfill
    \begin{minipage}[c]{\linewidth}
        \hfill
        \begin{minipage}[c]{0.37\linewidth}
        \includegraphics[trim={0cm 4.25cm 0cm 0},clip,width = \textwidth]{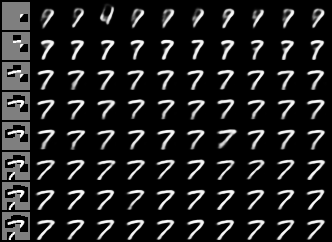}
        \centering{(a)}
        \end{minipage}
        \hfill
        \begin{minipage}[c]{0.37\linewidth}
        \includegraphics[trim={0cm 4.8cm 0cm 0},clip,width = \textwidth]{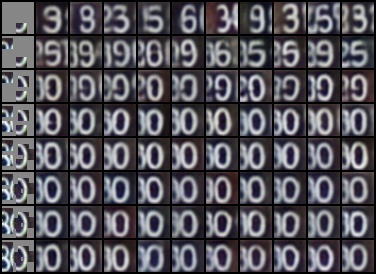}
        \centering{(b)}
        \end{minipage}
        \hfill
        \begin{minipage}[c]{0.235\linewidth}
        \includegraphics[width = \textwidth]{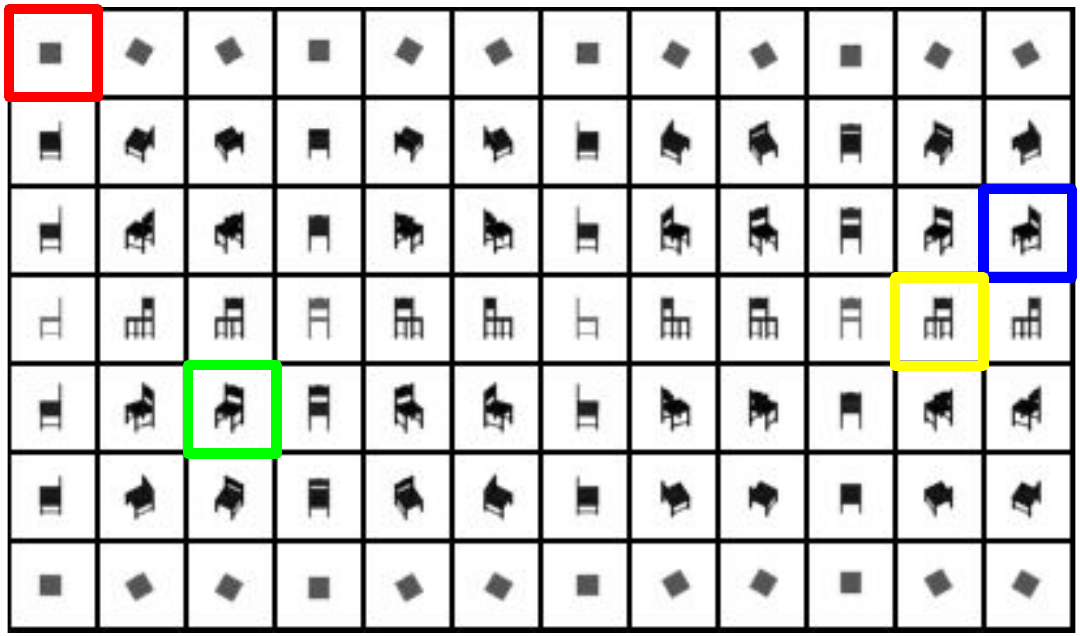}
        \centering{(c)}
        \end{minipage}
        \hfill
    \end{minipage}
    \vfill
    \begin{minipage}[c]{\linewidth}
        \includegraphics[width = \textwidth]{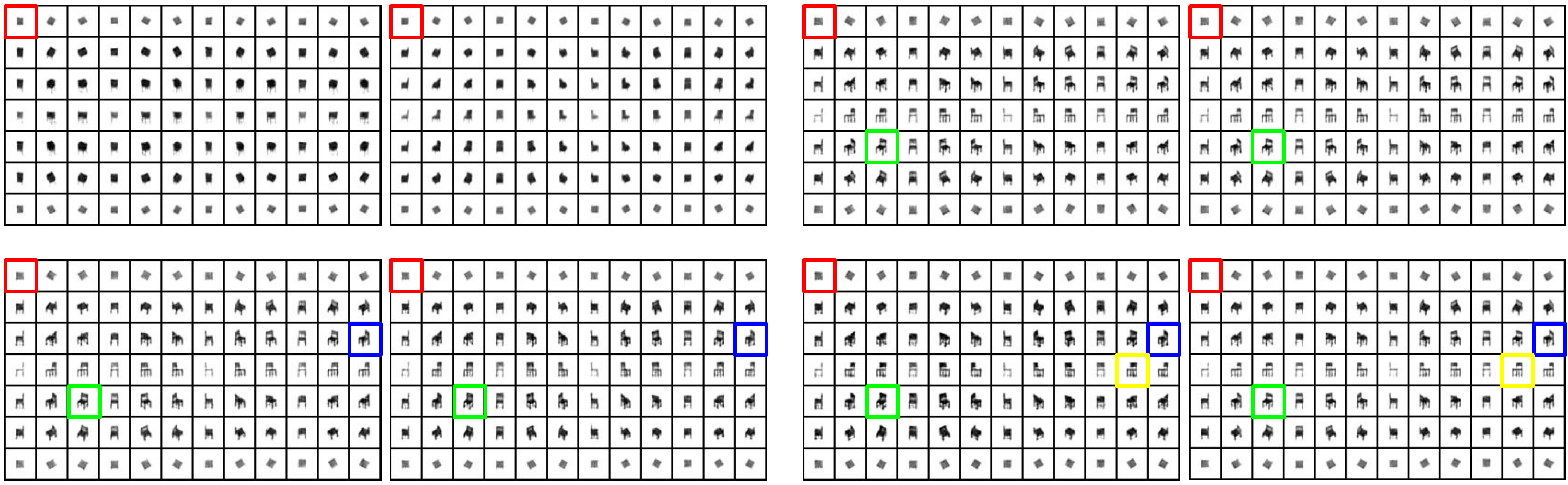}
        \centering{(d)}
    \end{minipage}
    \vfill
    \begin{minipage}[c]{\linewidth}
        \begin{minipage}[c]{0.16\linewidth}
        \vfill
            \begin{minipage}[c]{\linewidth}
                \includegraphics[width=\textwidth]{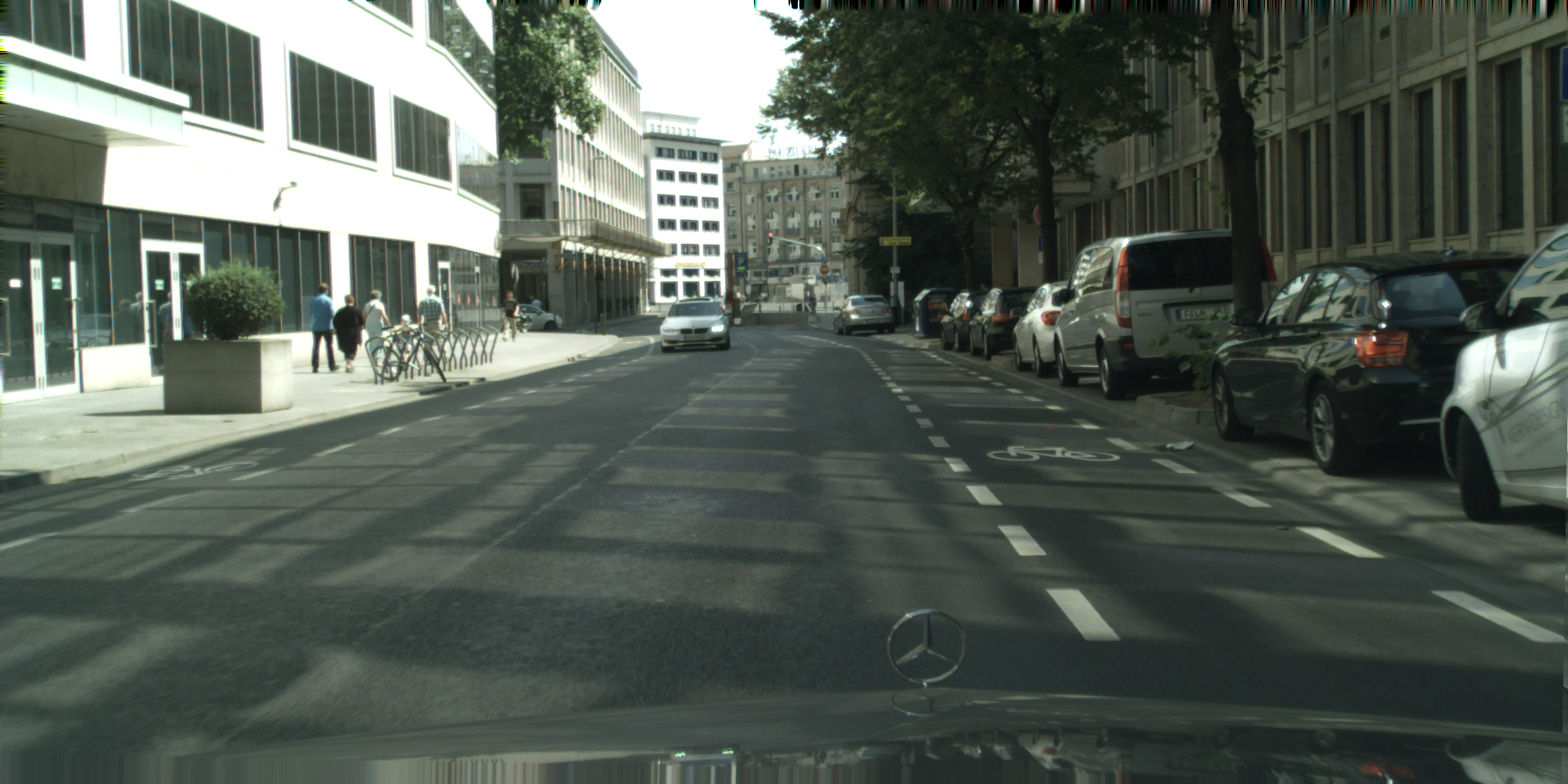}
                \centering{(e)}
            \end{minipage}
            \vfill
            \begin{minipage}[c]{\linewidth}
                \includegraphics[width=\textwidth]{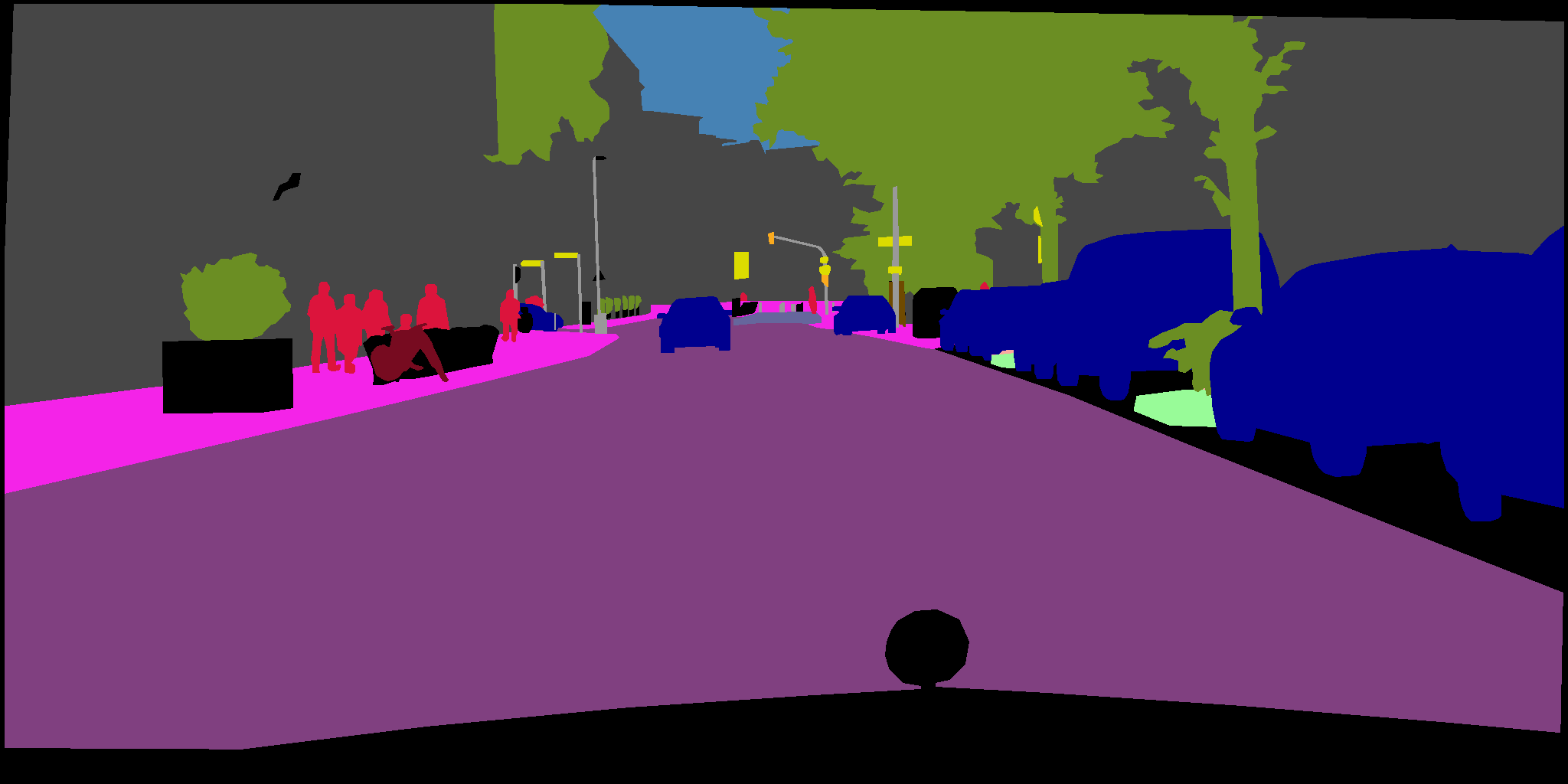}
                \centering{(f)}
            \end{minipage}
            \vfill
        \end{minipage}
        \hfill
        \begin{minipage}[c]{0.82\linewidth}
            \includegraphics[trim={0cm 0cm 22.5cm 0cm},clip,width = \textwidth]{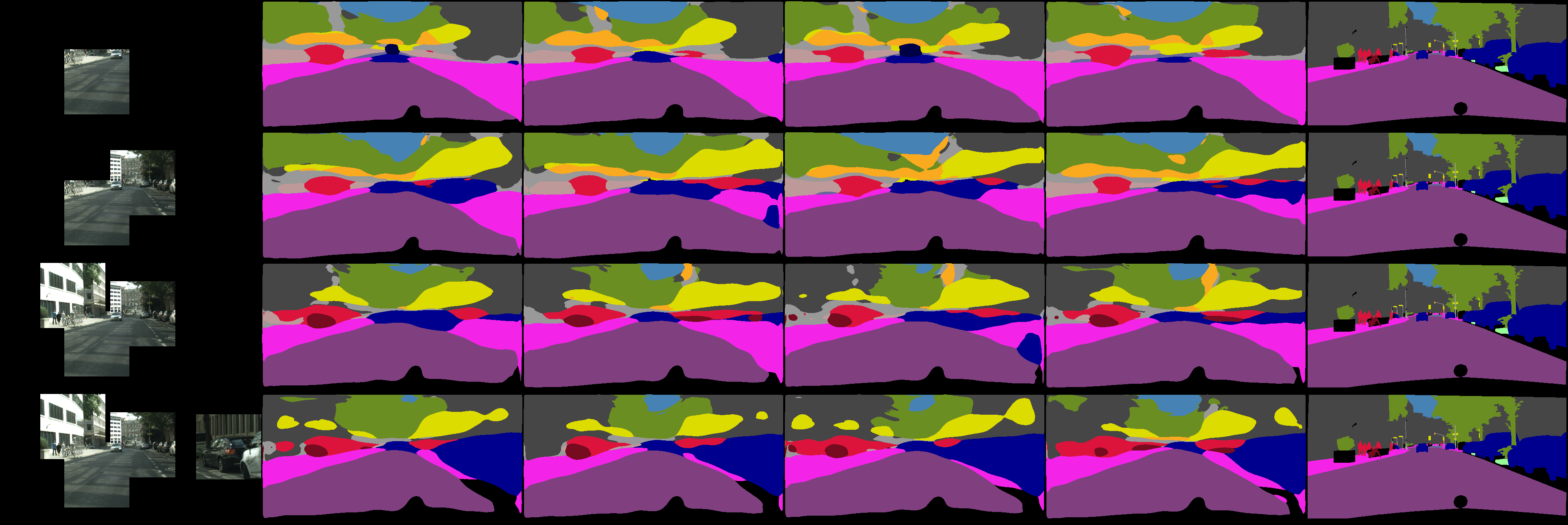}
            \centering{(g)}
        \end{minipage}
    \end{minipage}
    \vfill
    \caption{Samples of scenes predicted for 4 timesteps while observations are captured using uncertainty-based sensing. \textbf{(a-b)} Agents are trained on MNIST and SVHN. The first column shows the observation history, and the remaining columns show 10 prediction samples. Rows correspond to increasing timesteps. \textbf{(c)} An actual view-grid from ModelNet. An agent observes views highlighted in red, green, blue, and yellow from $t=1$ to 4. \textbf{(d)} Two samples of predicted view-grids from $t=1$ to 4 (left to right and top to bottom). \textbf{(e)} Actual scene from Cityscape explored by the agent in (g). \textbf{(f)} The ground-truth segmentation map of the scene. \textbf{(g)} An agent is exploring the scene shown in (e). The first column shows observation history and the remaining columns show 4 samples of predicted segmentation maps. Rows correspond to increasing timesteps. Zoom in to view details.}
    \label{fig:surroundingsamples}
\end{figure}

\subsection{Qualitative Analysis}
Predictions by the imaginative agents are shown in Fig.\ref{fig:surroundingsamples}. The agents trained on MNIST and SVHN (Fig.\ref{fig:surroundingsamples}(a-b)) predict different digits after looking at the first random glimpse. Predicted images have the same content at the fixated location. The successive glimpses are captured using the uncertainty-based sensing mechanism. The varieties in the predicted digits reduce as the agents receive further cues. The behavior of an agent trained on ModelNet is demonstrated on an example view-grid of a chair (Fig.\ref{fig:surroundingsamples}(c)). After looking at one random view, the agent predicts view-grids which seem to represent objects from the desk and the chair categories (Fig.\ref{fig:surroundingsamples}(d)). With the help of the second glimpse, the agent determines the global structure of the object. The following observations resolve uncertainty in the local details. For the cityscape dataset, an agent captures a new glimpse from an actual surrounding and hypothesizes complete segmentation maps. Instead of the variance-map, the next fixation is planned based on the pixel-wise entropy in the predicted segmentation maps. In Fig.\ref{fig:surroundingsamples}(g), the agent looks at a cycle-stand and a car in the first glimpse, and guesses a scene of an intersection. It guesses traffic lights and road signs on the intersection, and sidewalks and pedestrians on either side of the road. In the actual scene, there are cars parked on the right side of the road. Instead, the agent predicts a sidewalk, which is more likely in general. Through the following glimpses, the agent explores the configuration of the intersection point. It discovers a building on the left and the cars parked on the right. The agent strengthens it's belief and updates the imaginations with the next couple of glimpses. The segmentation maps computed using variational inference lack details but are useful for planning based on approximate locations of the various entities on the road.

\subsection{Quantitative Analysis}
The analysis is conducted using 1K, 1K, 100, and 50 predictions per-timestep for MNIST, SVHN, ModelNet, and Cityscape, respectively. The analysis is performed 10 times and the average results are presented next.

\begin{figure}[!t]
    \centering
    \begin{minipage}[c]{0.24\linewidth}
    \includegraphics[width = \textwidth]{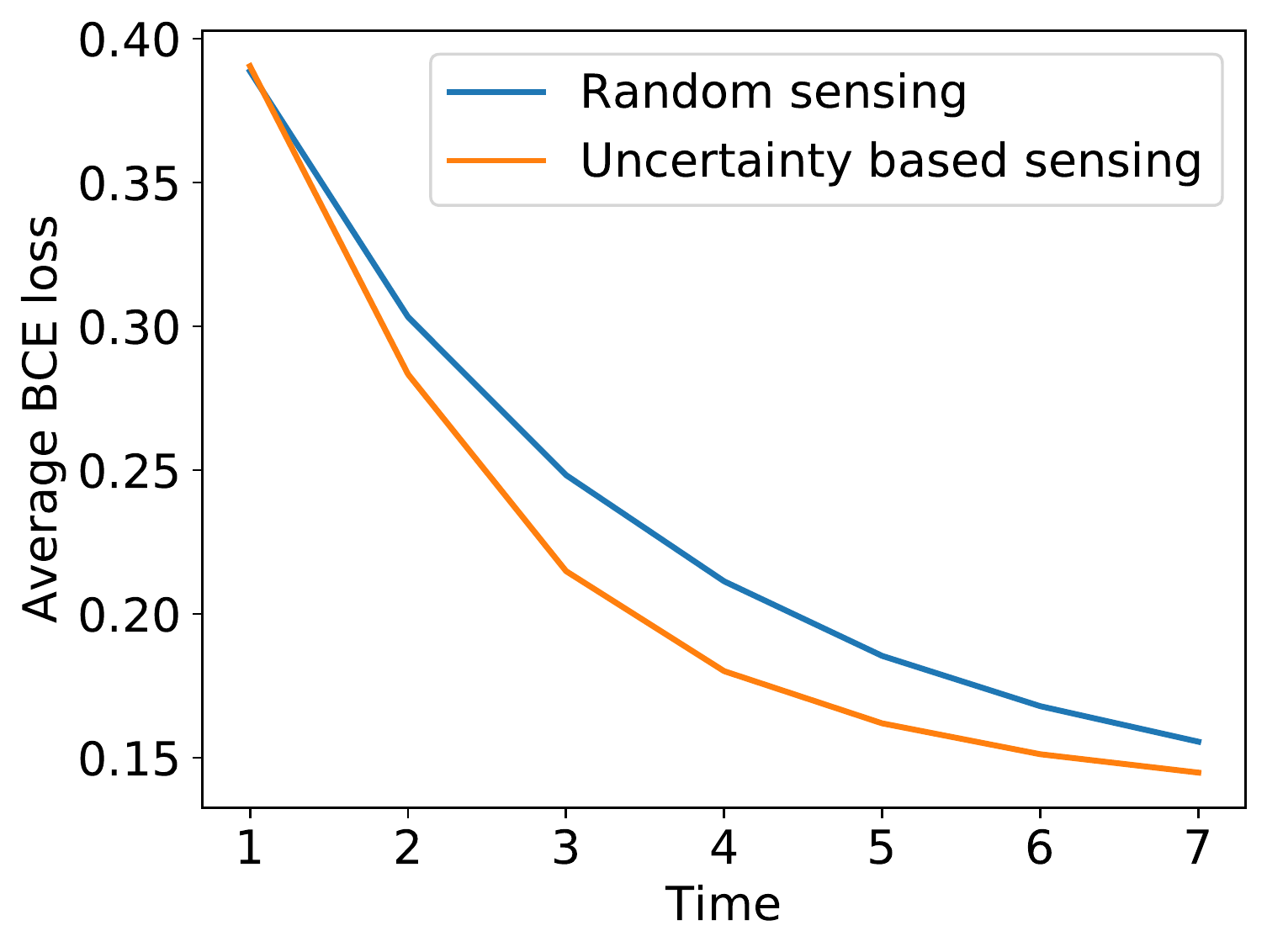}
    \centering{(a)}
    \end{minipage}
    \hfill
    \begin{minipage}[c]{0.24\linewidth}
    \includegraphics[width = \textwidth]{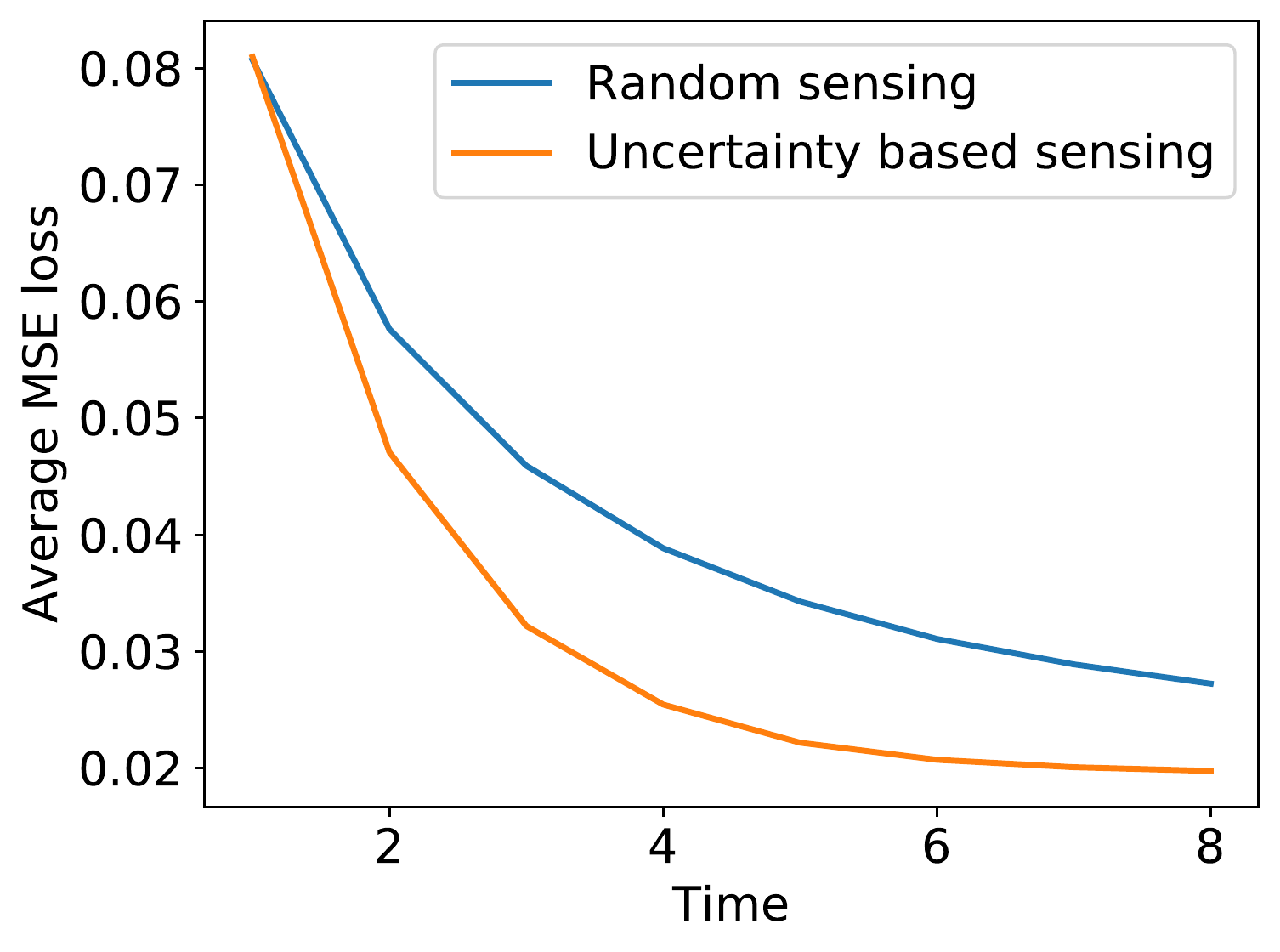}
    \centering{(b)}
    \end{minipage}
    \hfill
    \begin{minipage}[c]{0.24\linewidth}
    \includegraphics[width = \textwidth]{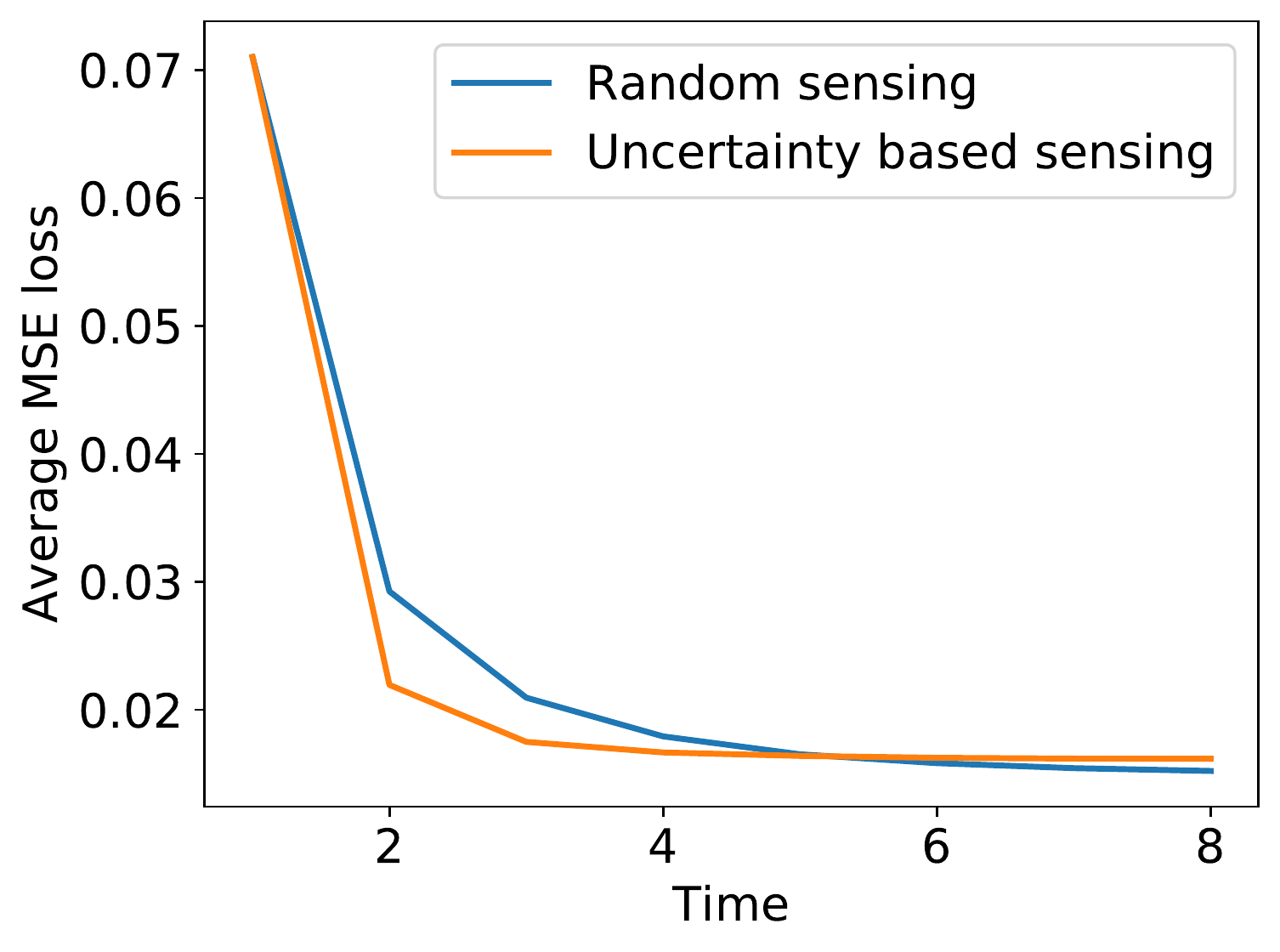}
    \centering{(c)}
    \end{minipage}
    \hfill
    \begin{minipage}[c]{0.24\linewidth}
    \includegraphics[width = \textwidth]{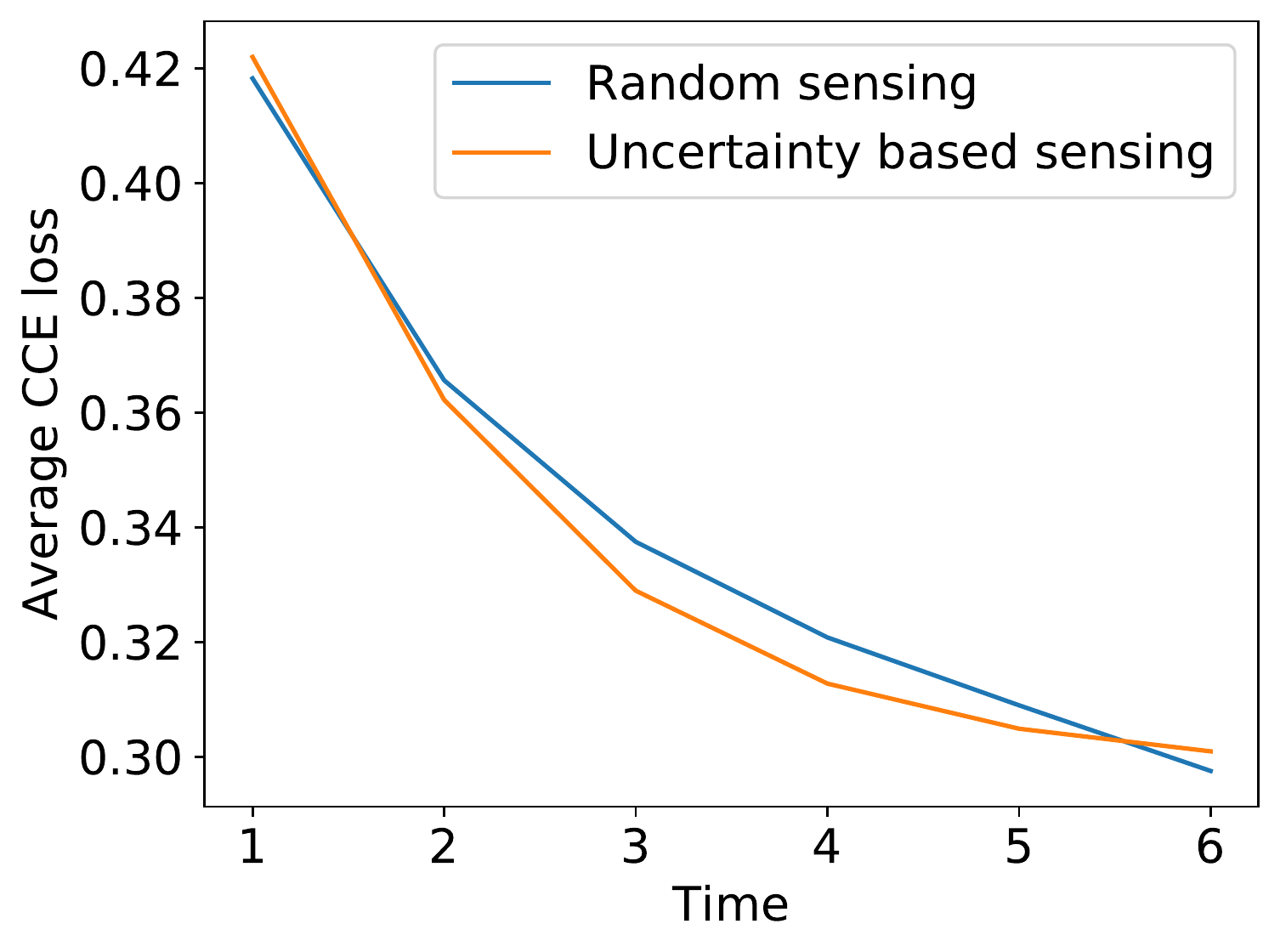}
    \centering{(c)}
    \end{minipage}
    \caption{Average of pixel-wise distances between actual scenes and the hypothesized scenes for increasing timesteps. Uncertainty-based sensing achieves better performance than random sensing. (a) MNIST (b) SVHN (c) ModelNet (d) Cityscape.}
    \label{fig:reconloss}
\end{figure}

\begin{figure}[!t]
    \centering
    \begin{minipage}[c]{0.24\linewidth}
    \includegraphics[width = \textwidth]{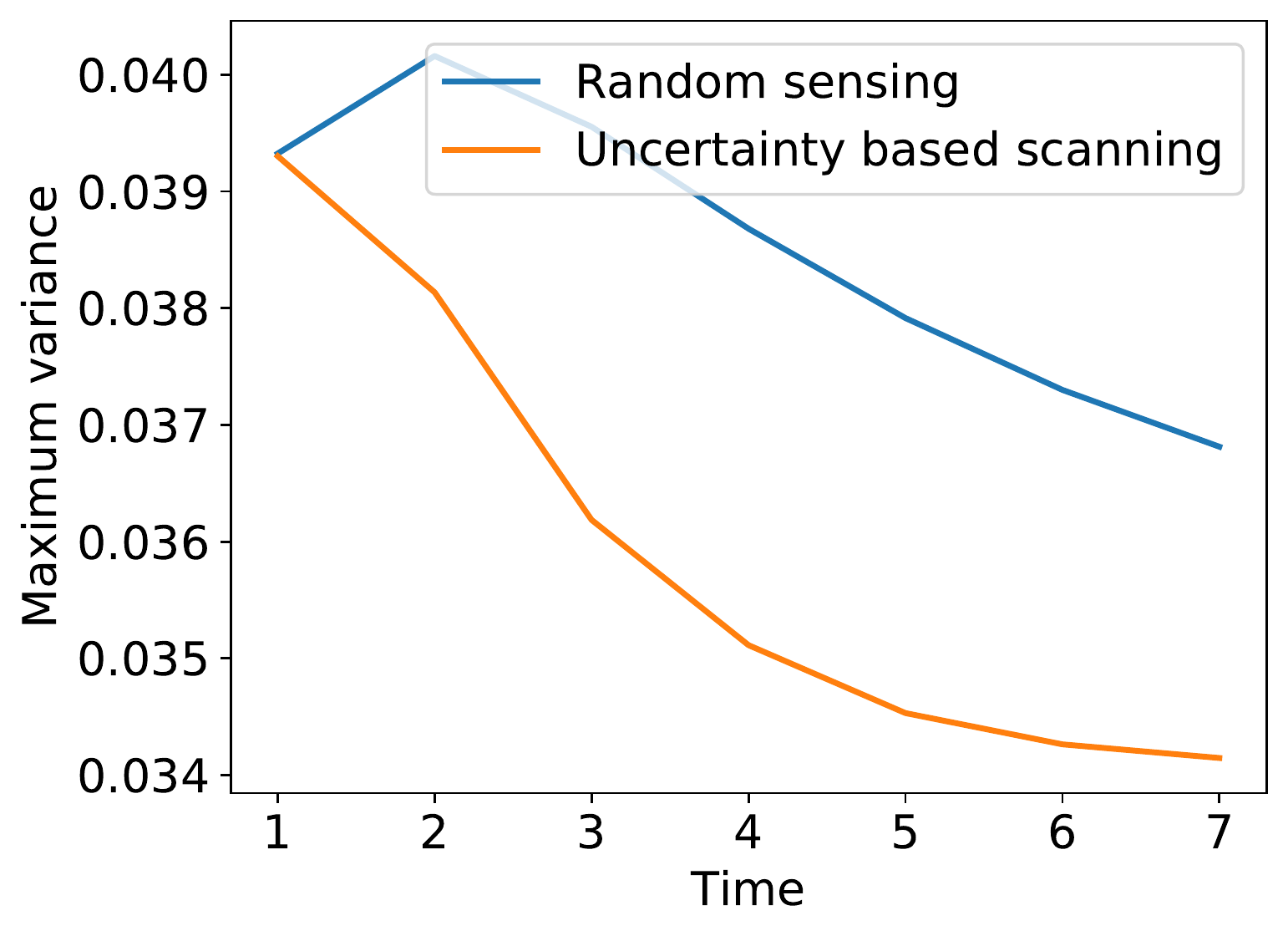}
    \centering{(a)}
    \end{minipage}
    \hfill
    \begin{minipage}[c]{0.24\linewidth}
    \includegraphics[width = \textwidth]{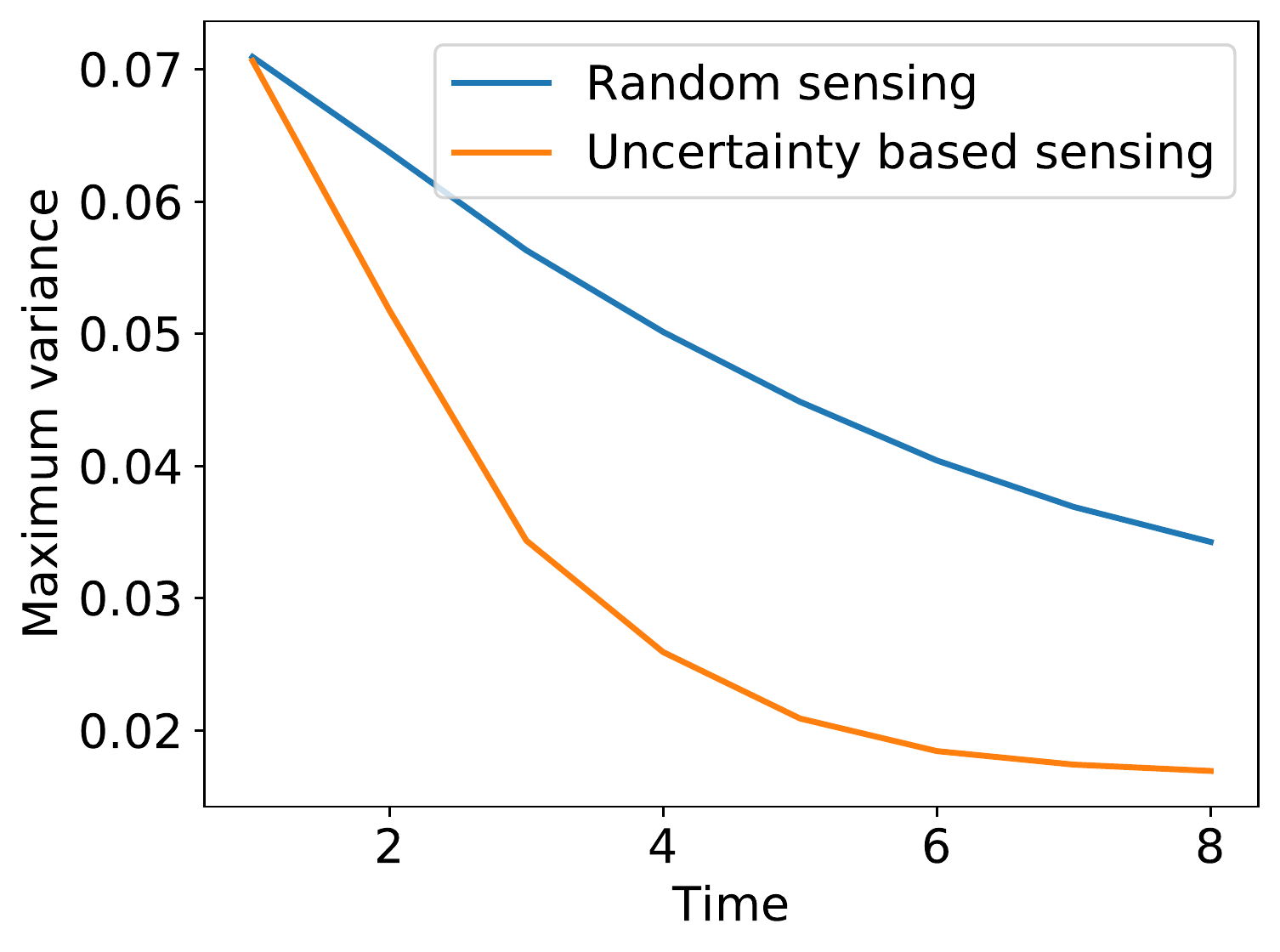}
    \centering{(b)}
    \end{minipage}
    \hfill
    \begin{minipage}[c]{0.24\linewidth}
    \includegraphics[width = \textwidth]{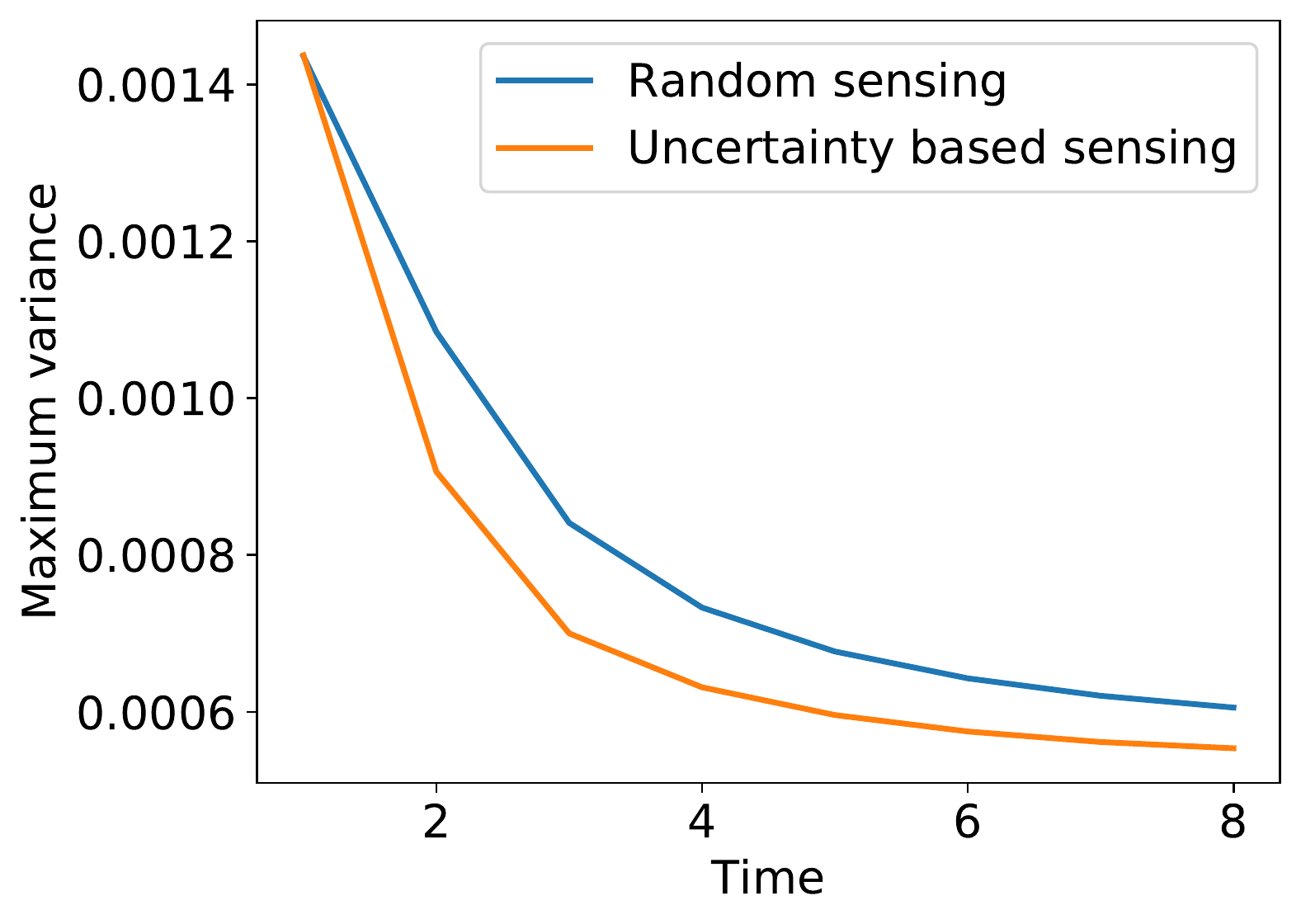}
    \centering{(c)}
    \end{minipage}
    \hfill
    \begin{minipage}[c]{0.24\linewidth}
    \includegraphics[width = \textwidth]{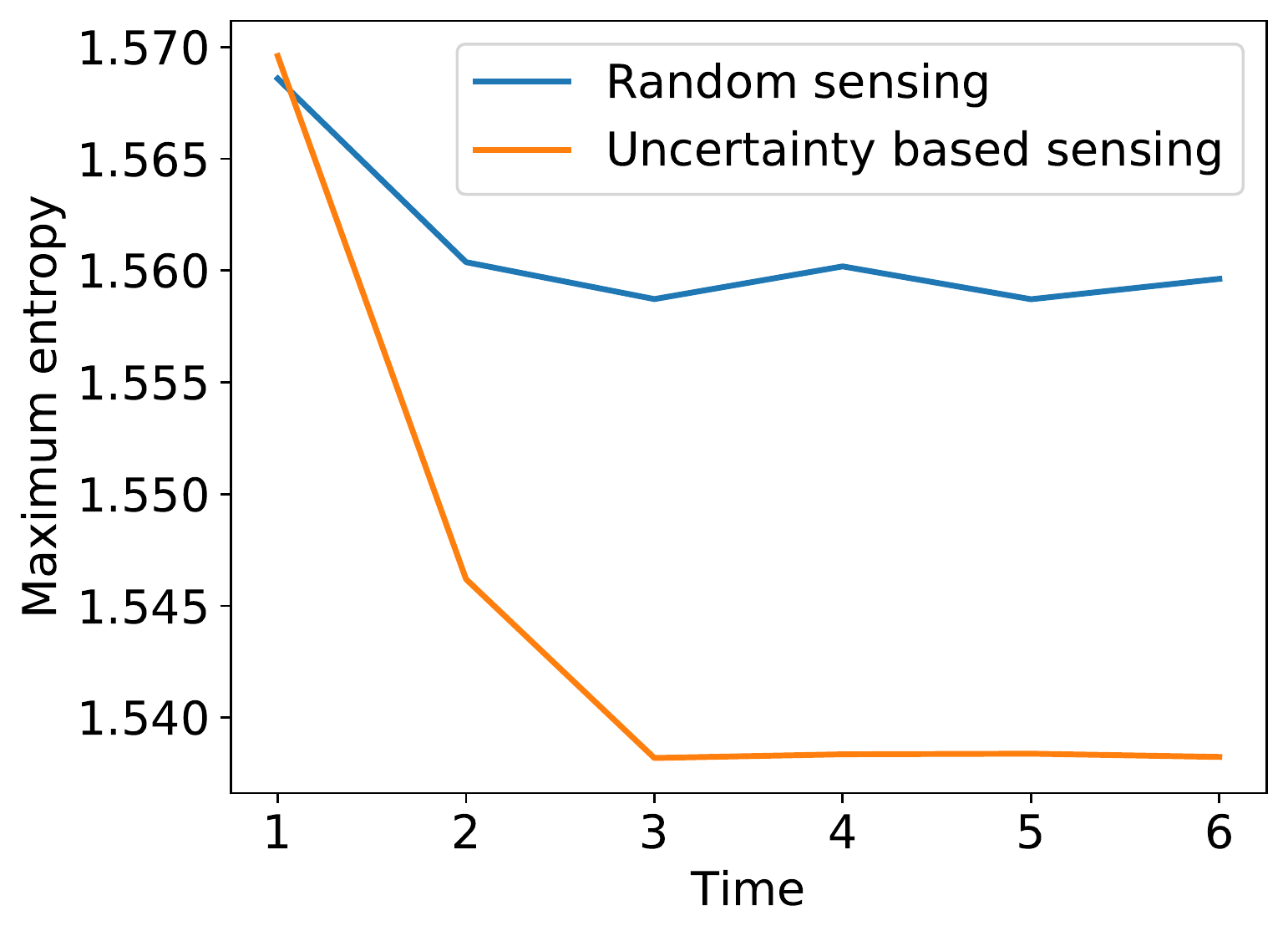}
    \centering{(c)}
    \end{minipage}
    \caption{Maximum values in the variance and entropy maps for increasing timesteps. The variance and the entropy reduces faster using uncertainty-based sensing compared to random sensing. (a) MNIST (b) SVHN (c) ModelNet (d) Cityscape.}
    \label{fig:maxvar}
\end{figure}

First, we compare the hypothesized scenes and an actual scene using an average of pixel-wise distances computed for increasing timesteps. The distance metrics used for comparison are binary cross-entropy for MNIST, mean squared error for SVHN and ModelNet, and categorical cross-entropy with median frequency balancing \cite{segnet} for Cityscape. The decreasing trends in Fig.\ref{fig:reconloss} suggest that the predicted scenes converge to the actual scene as time progresses. Also, the rate of convergence is higher with uncertainty-based sensing than sensing at random locations. The gain achieved using uncertainty-based sensing is lower for ModelNet and Cityscape as it is easy to determine the global structure of the scene from the very first glimpse. The remaining glimpses only resolve uncertainty in the local details. We also analyze how quickly an agent can reduce uncertainty in the content of the predicted scenes. As shown in Fig.\ref{fig:maxvar}, maximum values in the variance and entropy maps fall with time. The variety in the prediction reduces rapidly with uncertainty-based sensing compared to random sensing.

Next, we train classifiers to predict classes of the scenes hypothesized by the agents trained on MNIST, SVHN, and ModelNet. The classifiers predict class-labels from the latent representations of the hypotheses. Note that we do not aim at achieving the state of the art results in image classification. The main goal of this analysis is to demonstrate the effectiveness of the learnt representations for a scene-level task.

\begin{figure}[!t]
    \centering
    \begin{minipage}[c]{0.24\linewidth}
    \includegraphics[width = \textwidth]{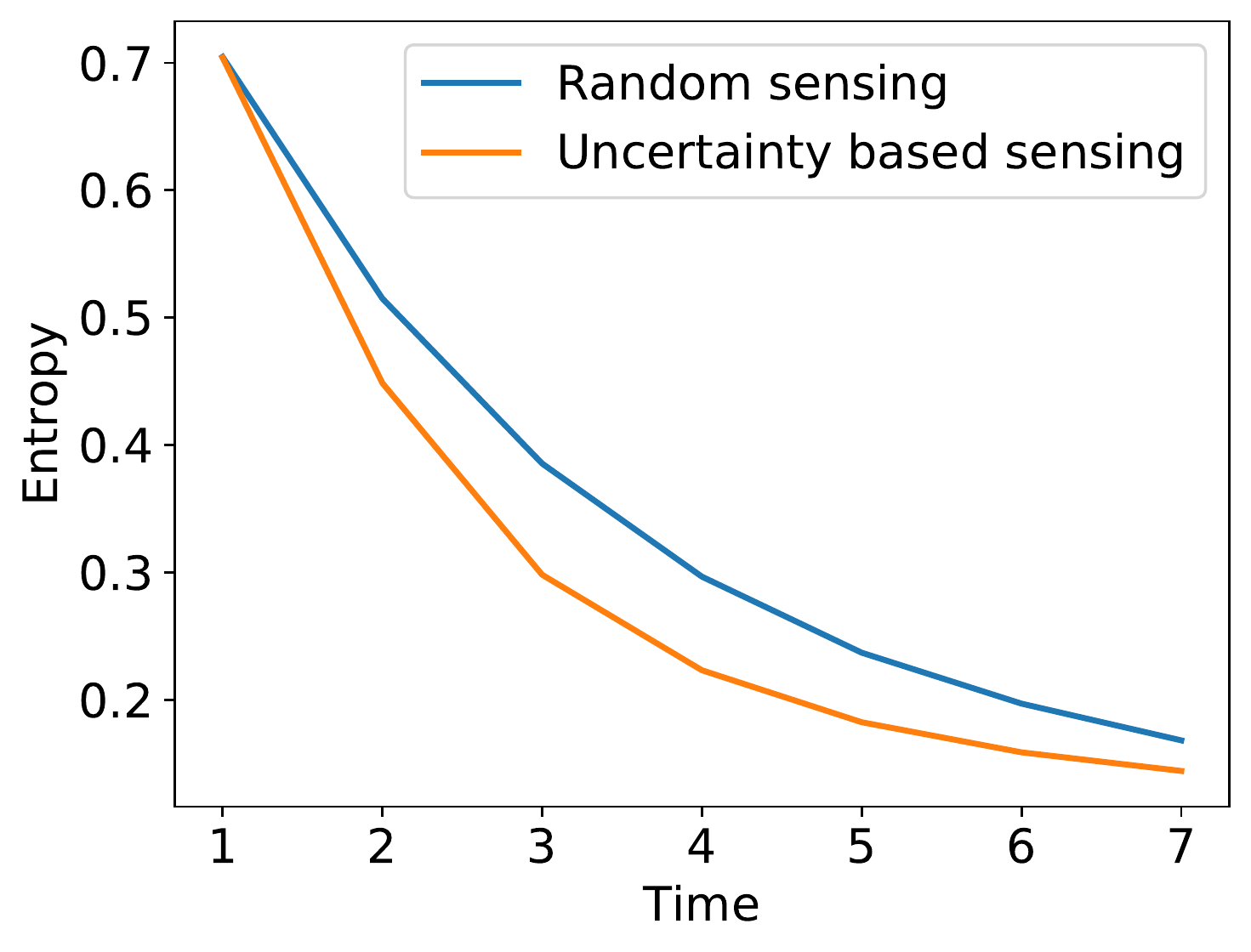}
    \centering{(a)}
    \end{minipage}
    \hfill
    \begin{minipage}[c]{0.24\linewidth}
    \includegraphics[width = \textwidth]{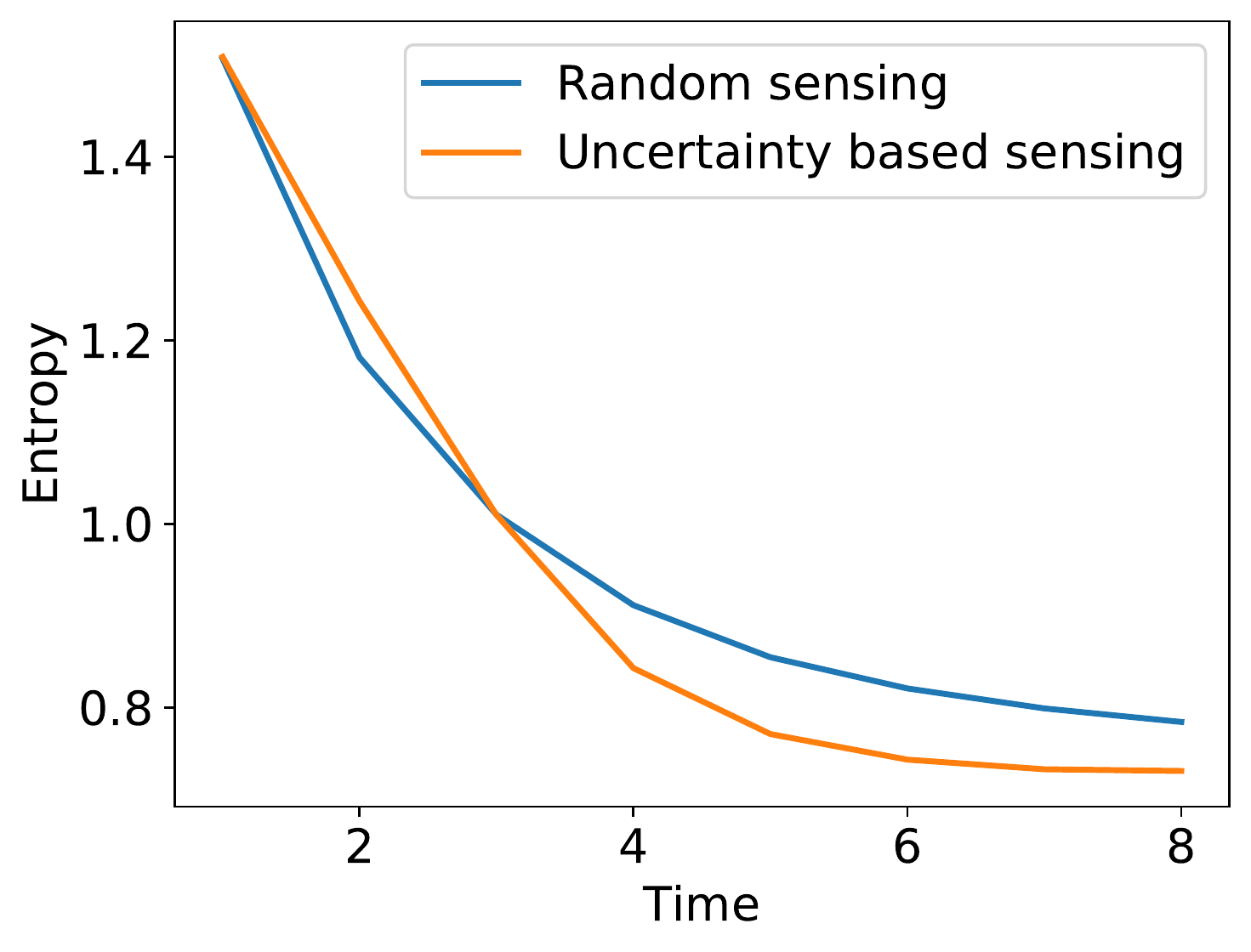}
    \centering{(b)}
    \end{minipage}
    \hfill
    \begin{minipage}[c]{0.24\linewidth}
    \includegraphics[width = \textwidth]{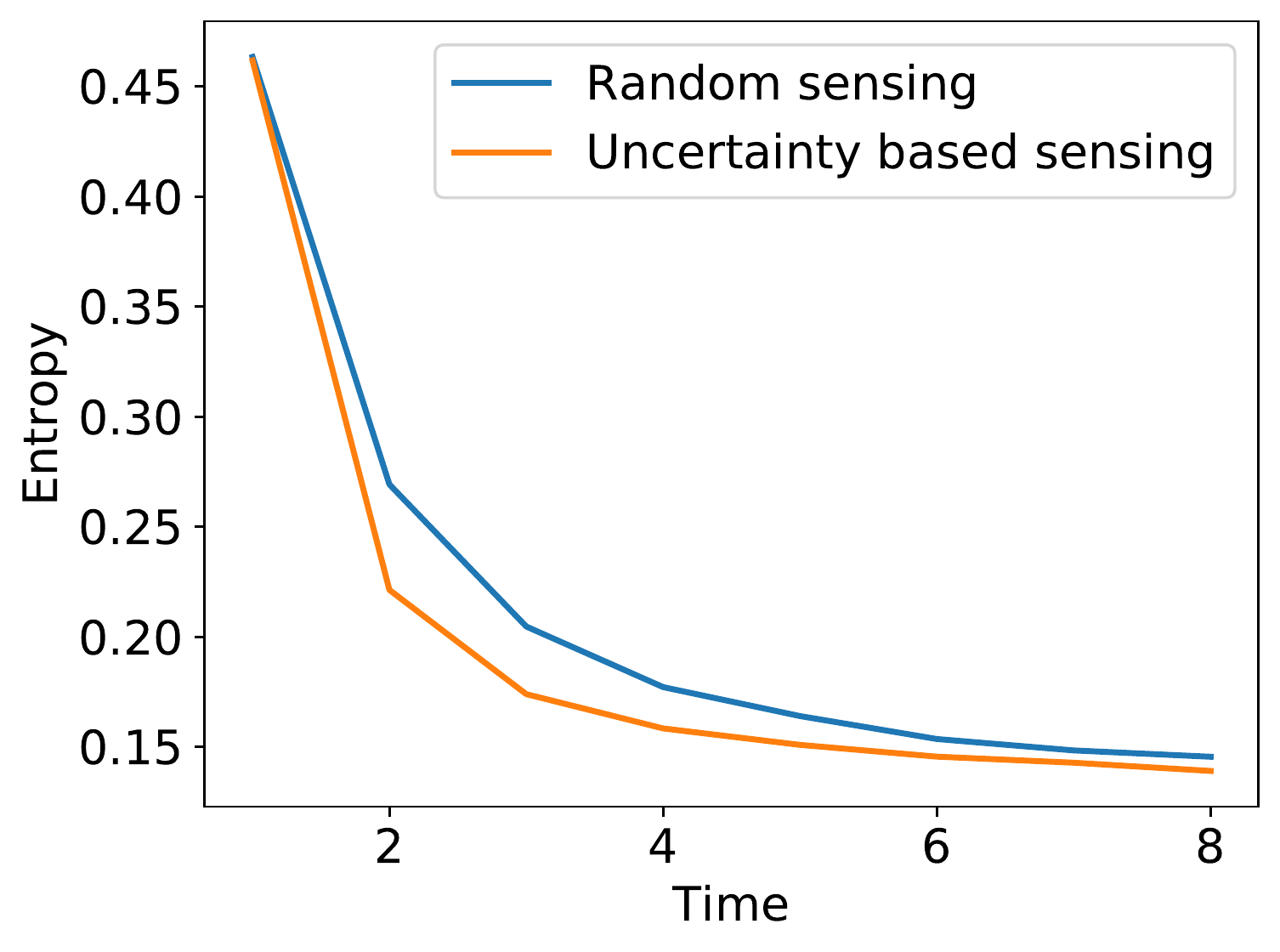}
    \centering{(c)}
    \end{minipage}
    \hfill
    \begin{minipage}[c]{0.24\linewidth}
    \includegraphics[width = \textwidth]{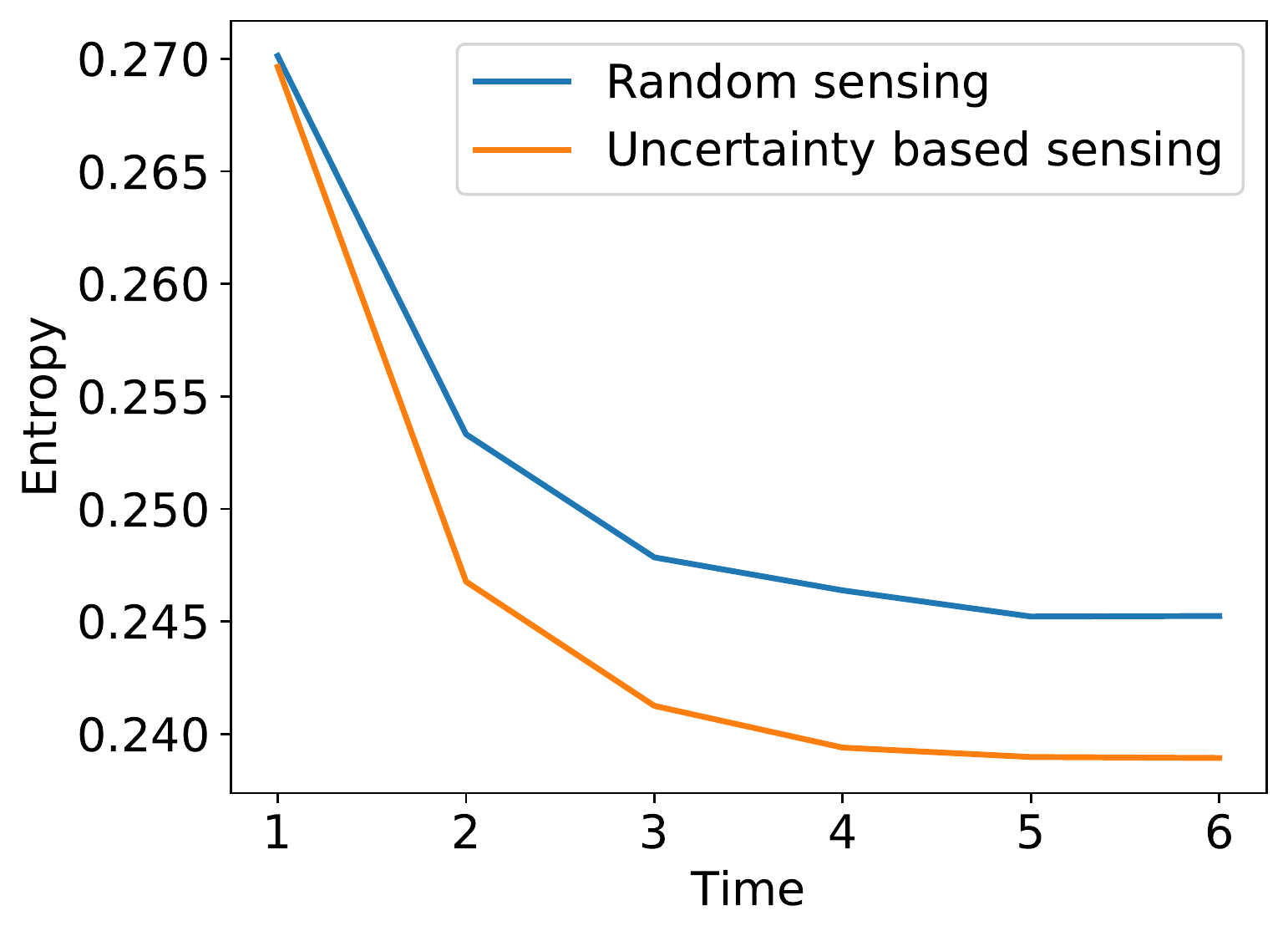}
    \centering{(d)}
    \end{minipage}
    \caption{Entropy in the categories of the scenes predicted at increasing timesteps. Uncertainty-based sensing reduces entropy faster compared to random sensing. (a) MNIST (b) SVHN (c) ModelNet (d) Cityscape.}
    \label{fig:probvsdec}
\end{figure}

\begin{figure}[!t]
    \centering
    \begin{minipage}[c]{0.24\linewidth}
    \includegraphics[width = \textwidth]{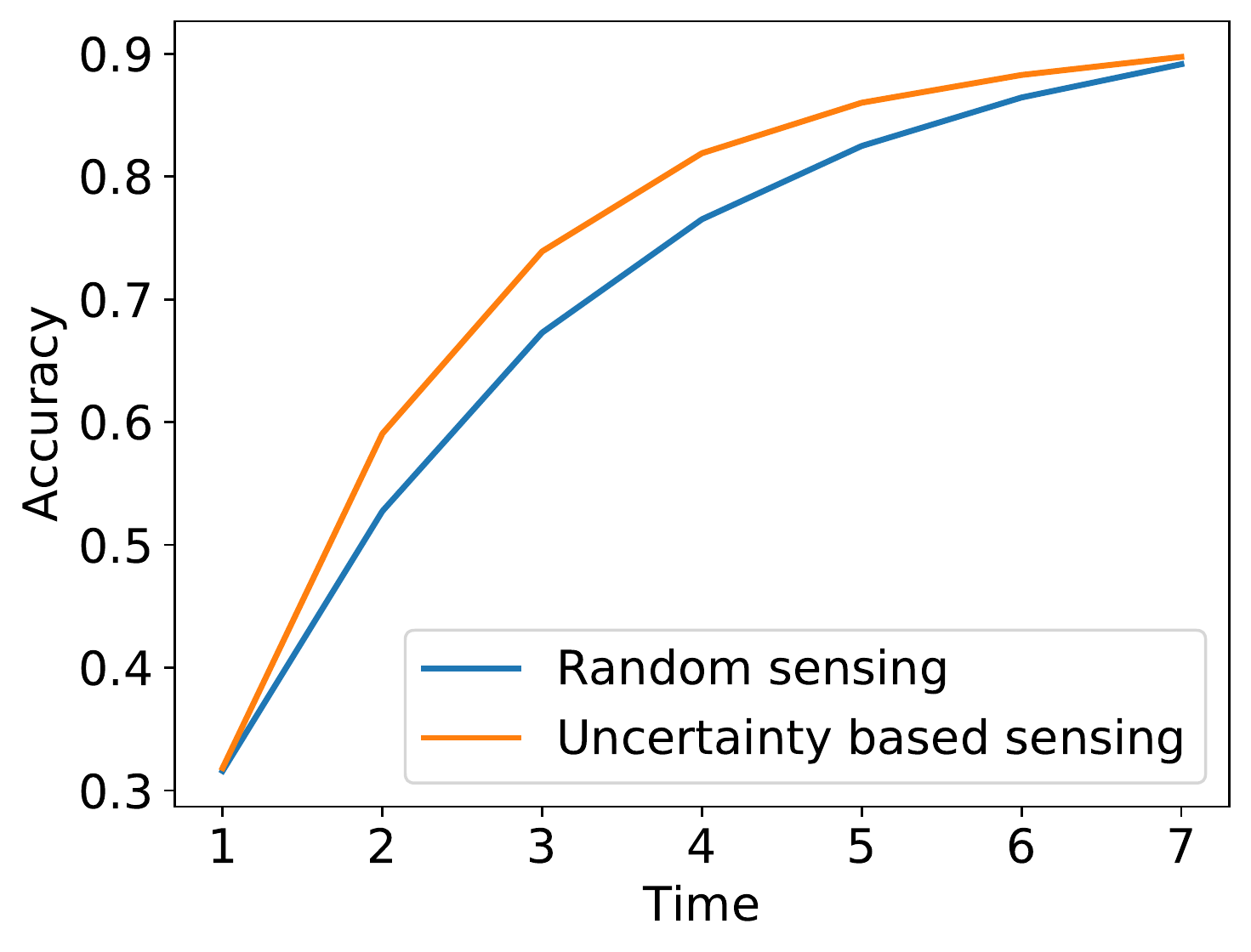}
    \centering{(a)}
    \end{minipage}
    \hfill
    \begin{minipage}[c]{0.24\linewidth}
    \includegraphics[width = \textwidth]{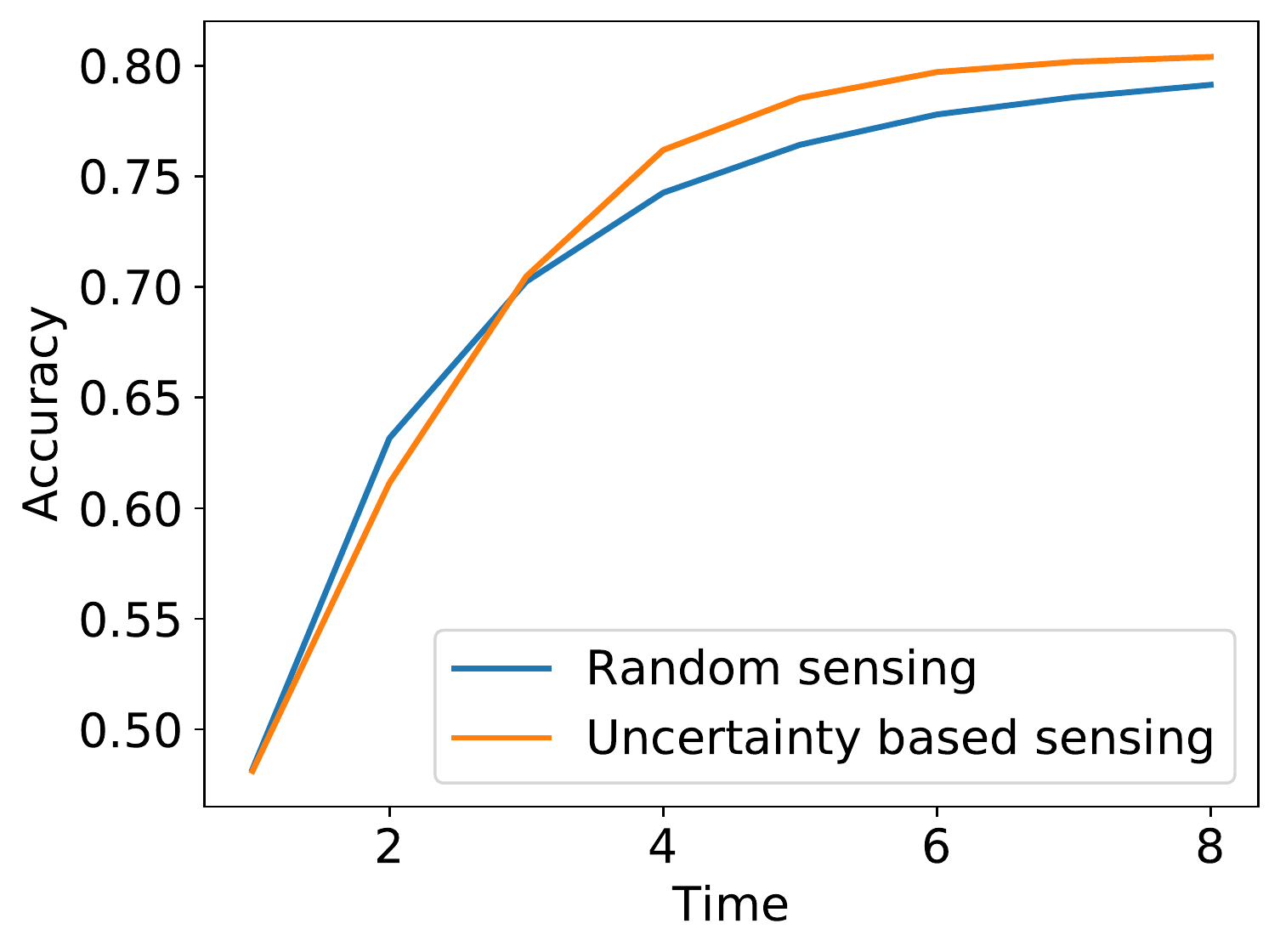}
    \centering{(b)}
    \end{minipage}
    \hfill
    \begin{minipage}[c]{0.24\linewidth}
    \includegraphics[width = \textwidth]{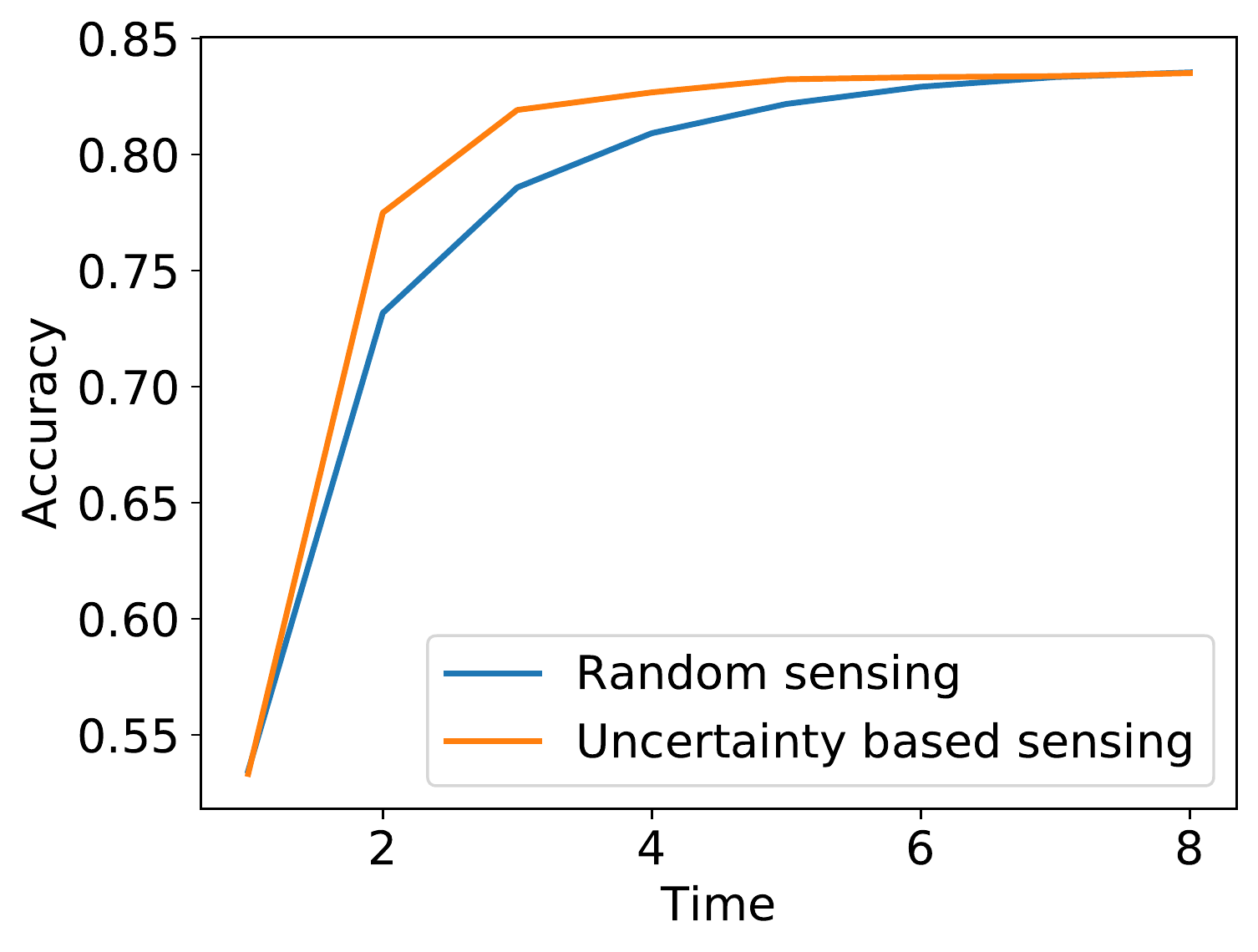}
    \centering{(c)}
    \end{minipage}
    \hfill
    \begin{minipage}[c]{0.24\linewidth}
    \includegraphics[width = \textwidth]{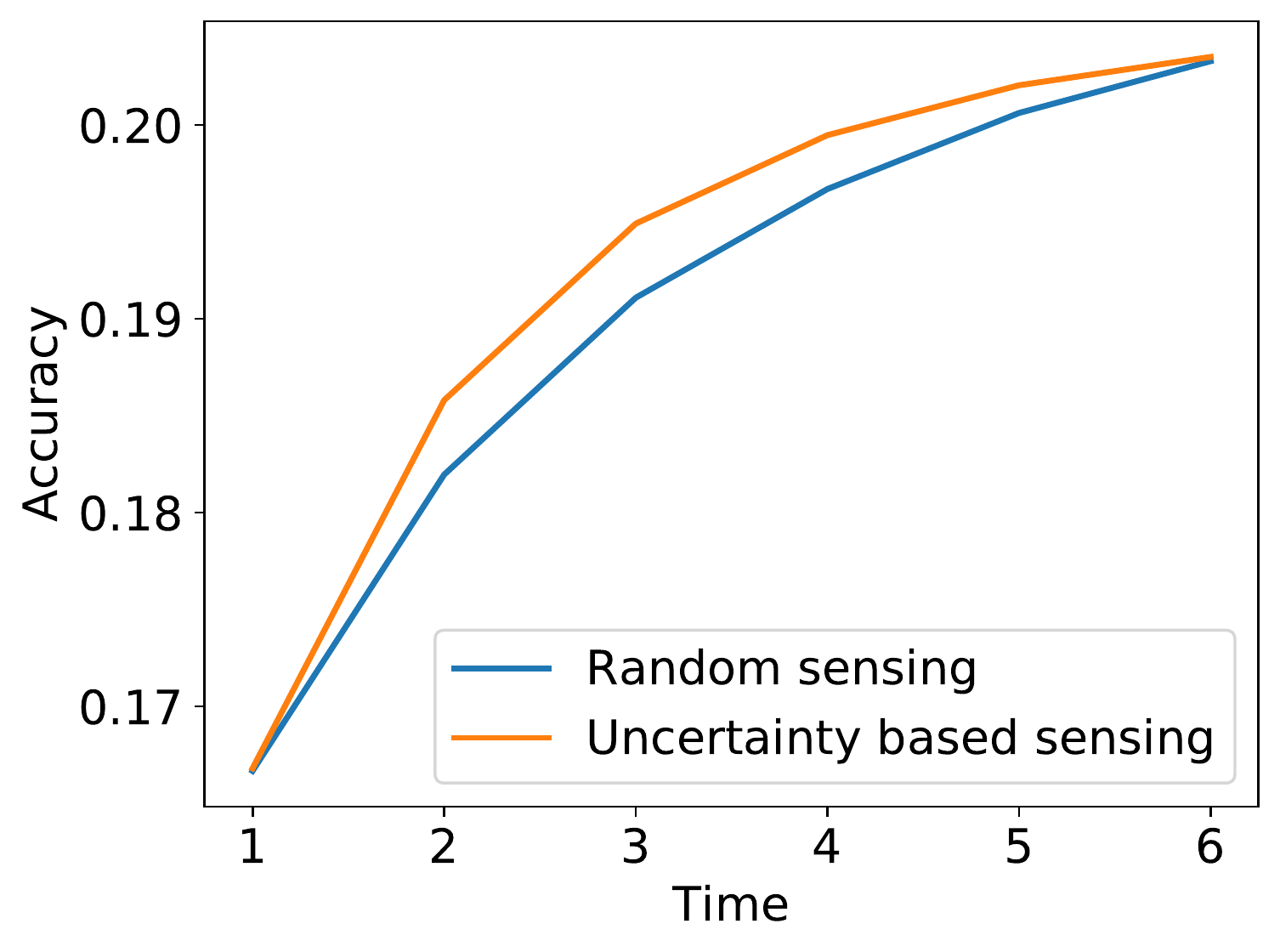}
    \centering{(d)}
    \end{minipage}
    \caption{Classification accuracy achieved at increasing timesteps. Uncertainty-based sensing outperforms random sensing. (a) MNIST (b) SVHN (c) ModelNet (d) Cityscape.}
    \label{fig:accvstime}
\end{figure}

We can analyze how quickly an agent concludes on the category of the scene. The agent may predict scenes from different categories provided that the content on the observed locations is consistent with the observations. As time progresses, the agent eliminates unlikely hypotheses, and the entropy in the categories of the predicted scenes reduces (see Fig.\ref{fig:probvsdec}(a-c)). Again, uncertainty-based sensing outperforms random sensing. We also analyze how quickly the category of an actual scene is identified. Fig.\ref{fig:accvstime}(a-c) show increasing trend in the classification accuracy. Higher performance can be achieved with a small number of glimpses captured using uncertainty-based sensing. The performance of the uncertainty-based sensing and the random sensing aligns when nearly the entire scene has been seen. Similar trends can be observed for entropy and average class accuracy of segmentation maps predicted for Cityscape (see Fig.\ref{fig:probvsdec}(d) and Fig.\ref{fig:accvstime}(d)). The class accuracy is the mean of the per-class accuracy.
\subsection{Comparison to Related Work}
\citet{lookaround,lookaround4} also proposed methods for unsupervised exploration of the scene. They hypothesize a single scene from the observation history. The predicted scene is blurry and non-realistic due to uncertainty. \citet{lookaround2} use a generative adversarial network \cite{gan} to make the predicted scene realistic. However, this approach still chooses a single hypothesis, which may not replicate the actual scene. We resolve the uncertainty by predicting multiple realistic scenes. The predictions and their latent representations can be used for planning by the higher-order processes. Additionally, our exploration policy is defined using the uncertainty in the predicted content. Unlike the previous works, the presented scheme is interpretable and does not require policy-learning.

A limitation of the proposed method is that it requires large datasets to learn posterior and likelihood densities accurately. Furthermore, optimization of the existing flow architectures become difficult for high dimensional latent spaces. Consequently, a laborious training procedure is required for large panoramic scenes.

\section{Conclusion}
We proposed a recurrent visual agent which perceives its surroundings through a series of partial observations. The imaginative agent hypothesizes unobserved portions of the scene and plans the next fixation using uncertainty in the hypotheses. The uncertainty in the predicted content reduces as the agent gathers additional cues. The uncertainty-based sensing helps in determining the actual surroundings faster than random sensing. The agent is probabilistic and is designed using variational autoencoders and normalizing flows. The presented architecture can be interpreted as conditional and unconditional VAE simultaneously. The internal representations of the predicted scenes are found to be useful for performing scene-level (e.g., classification) and pixel-level (e.g., segmentation) tasks. The agent is tested on multiple datasets with 2D and 3D scenes. The results display great potential in imagination-based attention. 

\clearpage

\bibliography{eccv2020submission}
\end{document}